%% file: template.tex
\documentclass[preprint,journal]{vgtc}            % preprint (journal style)

\onlineid{1547}

\ieeedoi{10.1109/TVCG.2024.3456305}

\vgtccategory{Research}

\vgtcpapertype{evaluation}

\title{Beware of Validation by Eye:\\ Visual Validation of Linear Trends in Scatterplots}

\author{%
  \authororcid{Daniel\ Braun}{0000-0002-8824-7184},
  \authororcid{Remco\ Chang}{0000-0002-6484-6430}, 
  \authororcid{Michael\ Gleicher}{0000-0003-3295-4071}, and 
  \authororcid{Tatiana\ von\ Landesberger}{0000-0002-5279-1444}
}

\authorfooter{
  \item
  	Daniel Braun, and Tatiana von Landesberger are with University of Cologne.
  	E-mail: \{braun\,$|$\,landesberger\}@cs.uni-koeln.de.
  \item
  	Remco Chang is with Tufts University.
  	E-mail: remco.chang@tufts.edu.
   \item
  	Michael Gleicher is with University of Wisconsin - Madison.
  	E-mail: gleicher@cs.wisc.edu.
}

\abstract{%
    \input{text/sec-abstract}  
}


\teaser{
  \centering
    \begin{subfigure}[t]{0.24\textwidth}
         \centering
         \includegraphics[width=0.9\textwidth]{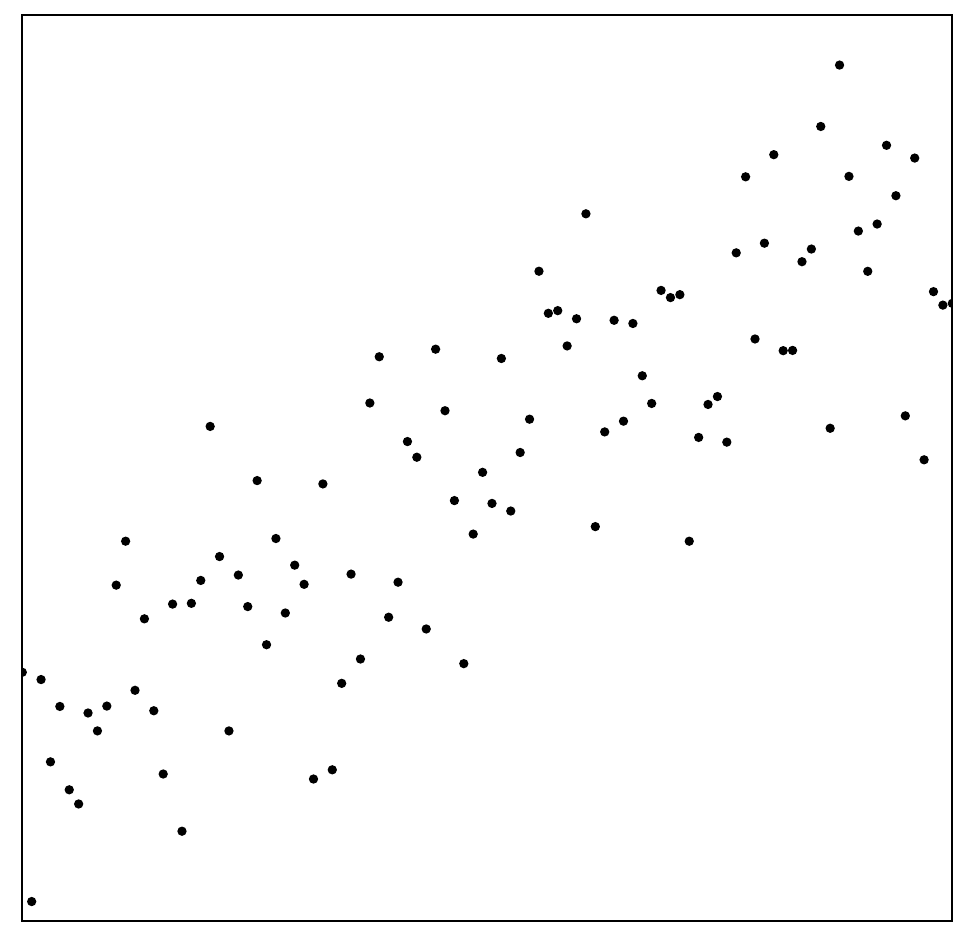}
         \caption{Scatterplot to visually estimate linear trend.}
         \label{fig:teaser_raw}
     \end{subfigure}
     \begin{subfigure}[t]{0.24\textwidth}
         \centering
         \includegraphics[width=0.9\textwidth]{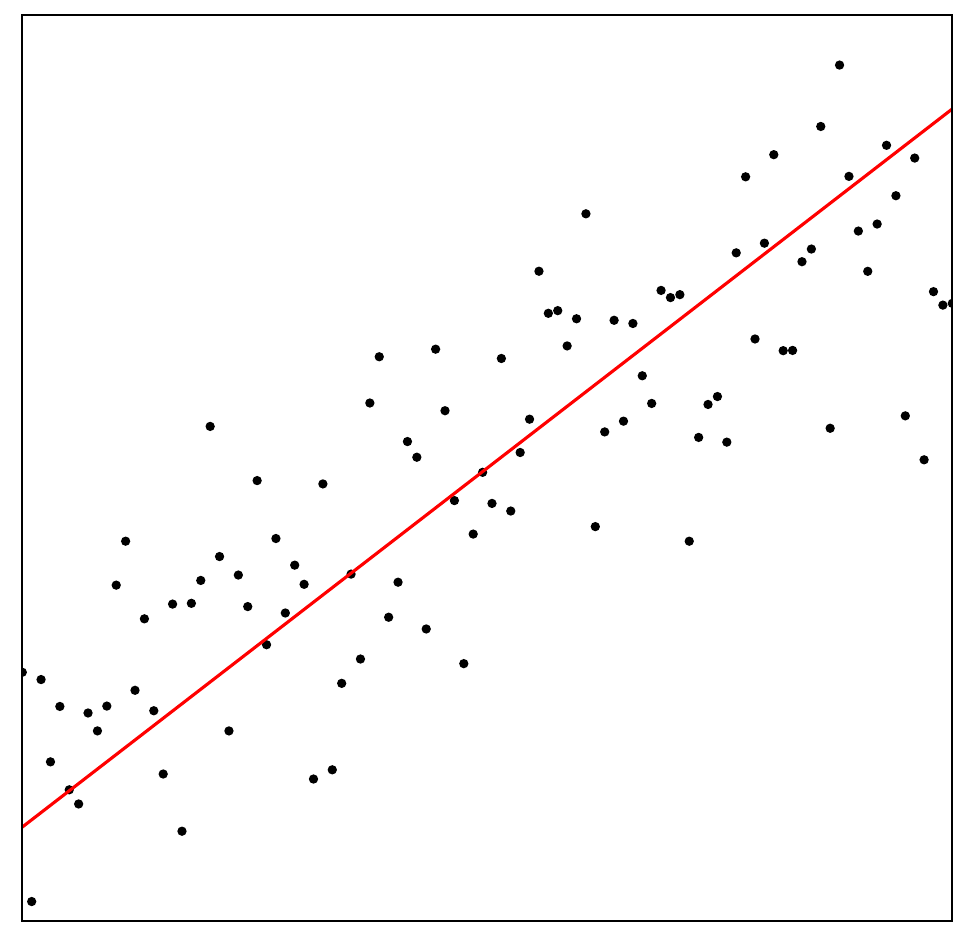}
         \caption{Trend line to be visually validated.}
         \label{fig:teaser_val}
     \end{subfigure}
     \hfill
     \begin{subfigure}[t]{0.24\textwidth}
         \centering
         \includegraphics[width=0.9\textwidth]{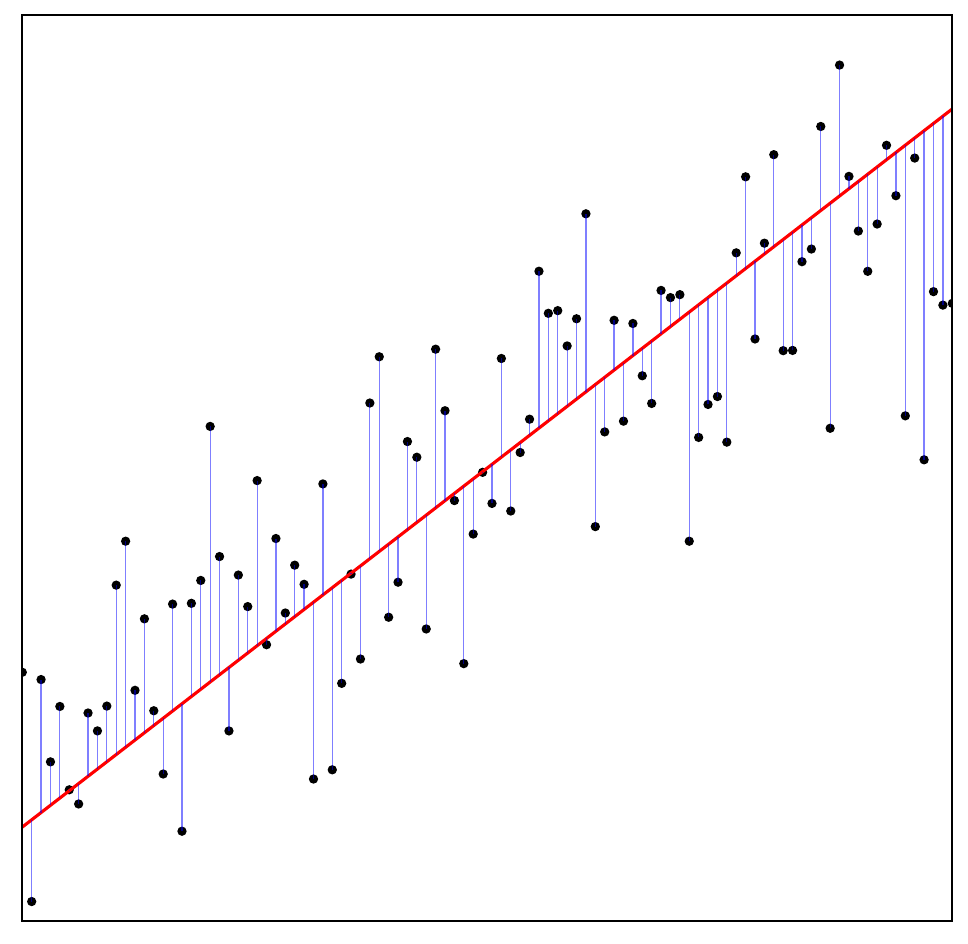}
         \caption{Visual design of adding error lines of ordinary least squares regression.}
         \label{fig:teaser_error}
     \end{subfigure}
     \hfill
     \begin{subfigure}[t]{0.24\textwidth}
         \centering
         \includegraphics[width=0.9\textwidth]{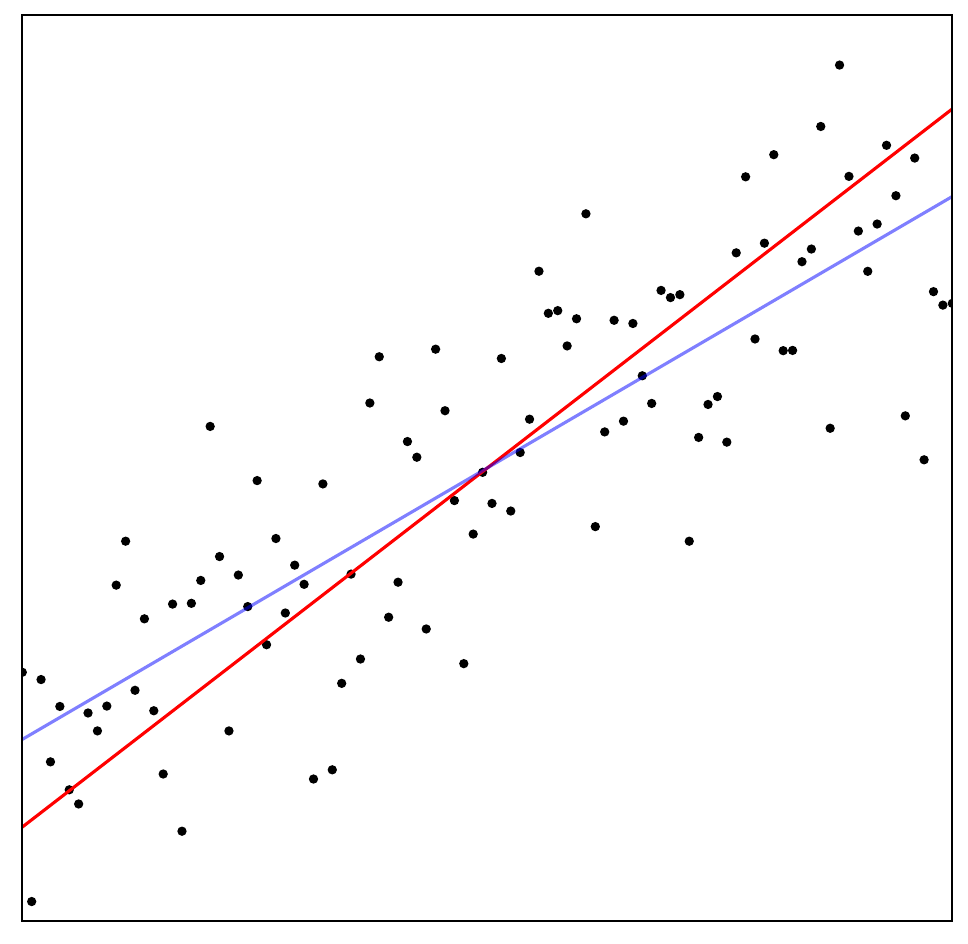}
         \caption{Example stimuli construction: True regression line (blue) and line to be validated (red, see \cref{fig:teaser_val}).}
     \end{subfigure}%
        \caption{Example scatterplot shown to participants in our user studies to investigate the visual validation of linear trends. In experiment~1 we tested the difference between visual validation (b) and visual estimation (a), while in experiment~2 (c) we compared different visual designs for regression to support visual validation. In (d), the blue line shows the true OLS regression.}
        \label{fig:teaser}
}

\graphicspath{{figs/}{figures/}{pictures/}{images/}{./}} % where to search for the images

\usepackage{tabu}                      % only used for the table example
\usepackage{booktabs}                  % only used for the table example
\usepackage{lipsum}                    % used to generate placeholder text
\usepackage{mwe}                       % used to generate placeholder figures

\usepackage{mathptmx}                  % use matching math font

\begin{document}

%%%%%%%%%%%%%%%%%%%%%%%%%%%%%%%%%%%%%%%%%%%%%%%%%%%%%%%%%%%%%%%%
%%%%%%%%%%%%%%%%%%%%%% START OF THE PAPER %%%%%%%%%%%%%%%%%%%%%%
%%%%%%%%%%%%%%%%%%%%%%%%%%%%%%%%%%%%%%%%%%%%%%%%%%%%%%%%%%%%%%%%

\firstsection{Introduction}

\maketitle

\input{text/sec-intro}

\input{text/sec-rw}

\input{text/sec-study}

\input{text/sec-exp1}

\input{text/sec-exp2}

\input{text/sec-futurework}

\input{text/sec-conclusion}

\section*{Supplemental Materials}
\label{sec:supplemental_materials}

Supplemental material includes: 1) Python code (Jupyter-notebook) for the data and visual designs generation; 2) User study documentations including the stimuli as PDF files; and 3) Anonymized study results as CSV files.

%% if specified like this the section will be omitted in review mode
\acknowledgments{%
	The authors would like to thank all study participants. This paper is a result of Dagstuhl Seminar 22331 ``Visualization and Decision Making Design Under Uncertainty''. This work has been partially supported by the BMBF WarmWorld Project, the SANE Project, and the Risk-Principe Project. This work has also been funded in part by NSF Awards 2007436, 1452977, and 2118201.%
}

\bibliographystyle{abbrv-doi-hyperref}

\bibliography{literature}

\end{document}

%% file: text/sec-abstract.tex
Visual validation of regression models in scatterplots is a common practice for assessing model quality, yet its efficacy remains unquantified. We conducted two empirical experiments to investigate individuals' ability to visually validate linear regression models (linear trends) and to examine the impact of common visualization designs on validation quality.
The first experiment showed that the level of accuracy for visual estimation of slope (i.e., fitting a line to data) is higher than for visual validation of slope (i.e., accepting a shown line). Notably, we found bias toward slopes that are ``too steep'' in both cases. This lead to novel insights that participants naturally assessed regression with orthogonal distances between the points and the line (i.e., ODR regression) rather than the common vertical distances (OLS regression).
In the second experiment, we investigated whether incorporating common designs for regression visualization (error lines, bounding boxes, and confidence intervals) would improve visual validation.
Even though error lines reduced validation bias, results failed to show the desired improvements in accuracy for any design.
Overall, our findings suggest caution in using visual model validation for linear trends in scatterplots.

%% file: text/sec-intro.tex
Visual validation of statistical models serves as an important task in modern statistics and machine learning applications. The complexity of models often demands visual inspection for assessing their correctness and reliability~\cite{Chatzimparmpas.2020, Chatzimparmpas.2020a}. This is crucial, as the model outcomes have great implications in critical domains~\cite{Hullman.2021}, e.g., the estimation of pandemic outbreaks~\cite{dansana2020global} and meteorological forecasting~\cite{Jeuken.1996}. Without visualization, traditional statistical metrics are often insufficient in describing the underlying data and model. For example, datasets with significantly different characteristics can share the same numerical metrics~\cite{Matejka.2017, Blyth.1972, Szafir.2016}. As a result, visualization researchers have advocated for visual validation of statistical models as a core part of data analysis. 

To date, most research on the perception of statistical models has focused on \textit{visual estimation} -- individuals' ability to visually fit a model to data~\cite{Correll.2017, Gleicher.2013, Harrison.2014, Hong.2022, Kay.2016, Newburger.2022, Rensink.2010, Xiong.2022, Yang.2019, Yuan.2019, Moritz.2024, Ciccione.2021, Ondov.2021, Strain.2023a, Strain.2023b, Reimann.2021}. While these studies contribute to our understanding of the visual estimation process, there is a lack of research on \textit{visual model validation} -- individuals' ability to assess the fit of a given model to the underlying data.
In a recent study by Braun et al.~\cite{Braun.2023}, the authors found significant differences in individuals' performance in visual validation and estimation of averages in scatterplots. 
In this paper, we build upon this work and investigate the accuracy and effectiveness of visual model validation for a more complex model -- \textit{linear trends} in scatterplots.
While the average value offers a fundamental measure in one dimensions, the exploration of linear trends provides valuable insights into the relationships between two data dimensions, which can be particularly relevant in machine learning applications where linearity assumptions are common.
Scatterplots are the standard approach for showing relationships between two quantitative variables~\cite{Munzner.2014}.

Linear trends can be interpreted in two different ways~\cite{Bewick.2003}:  \textit{Regression} and \textit{correlation}. Regression estimates the underlying linear relationship between two variables to find a function that predicts the value of one variable based on the other. Correlation measures the strength of the linear relationship between two variables. The closer points lie to a straight line, the stronger their relationship. Due to the relevance of regression models for decision making in wide-ranging fields~\cite{Dawes.1974, Dawes.1979, Sultan.2021, Morton.2003}, visual validation of regression models is important. In this paper, we investigate the perception of linear regression models and answer the following research questions:

\begin{itemize}[noitemsep, leftmargin=*]
    \item \textbf{RQ1}: How does performance in visual validation of linear trends relate to the accuracy of visual estimation?
    \item \textbf{RQ2}: Can common visual designs enhance the performance of visual validation of linear trends?
\end{itemize}

%\begin{figure}[htbp]
% \centering
% \includegraphics[width=\columnwidth]{trend}
% \caption{Comparing local trends with one global trend~\cite{Tableau}.
% \label{fig:tableau}
%\end{figure}

For RQ1, we examined individuals' perception of the slope of a linear trend in a between-subject user study. Participants were randomly assigned to either validate the slope of a shown trend line in a scatterplot (\cref{fig:teaser_val}) or estimate the trend line on their own (\cref{fig:teaser_raw}).

Our results confirm the previous findings of Braun et al.~\cite{Braun.2023} that participants are more accurate in model estimation than in model validation. Further, we found that participants systematically overestimated the trends' slope and were particularly inaccurate at recognizing trend lines with a high slope value (i.e., lines that are ``too steep''). This implies that the perception of the trend lines is biased -- not symmetrical, i.e., lines with slopes that are too high and too low are perceived differently. A post-hoc analysis revealed that participants' responses were more consistent with the non-standard orthogonal regression (ODR: minimizes orthogonal distances between data points and regression line) than with the usual vertical regression (ordinary least squares - OLS: minimizes vertical distances between data points and regression line), since ODR regression has a higher slope per calculation (see ~\cref{fig:ols_odr}). This implies they assumed errors in both x and y variables rather than just one. Our results show similar effects between positive and negative trends.

With respect to visual design, several recent studies have found significant effects of visualization on model estimation~\cite{Strain.2023a, Strain.2023b, Moritz.2024, Yang.2019, Szafir.2016, Harrison.2014}. We similarly investigate whether the addition of augmentations to the visual designs improve individuals' performance in the visual validation of models (RQ2). To answer this research question, we modified the visual designs and repeated our first study on the slope validation. Using the same data, we added three visual designs to the shown trend line - OLS error lines, 95\% confidence intervals, and bounding boxes (see ~\cref{fig:designs}).  
The results showed that the addition of error lines reduced the bias in recognizing lines with slopes that are either too high or too low. The addition of bounding boxes slightly increased visual model validation accuracy. However, none of the designs lead to the desired improvements (i.e, higher acceptance rate for correct models and higher rejection rate for incorrect models). In fact, participants reported an increased task difficulty with the addition of visual designs with no benefits to task completion time.

Altogether, the results of our two studies find evidence to caution when using the common practice of ``validation by eye''~\cite{Hullman.2021} -- visually validating statistical models by overlaying a visualization of the model over raw data. Further research is needed on how to support individuals in the visual validation process.
%Our first study result strongly suggests that participants are more accurate when estimating a regression model in a scatterplot than validating a given regression model result. 
%Our second study finds that, although different commonly used visualization designs can reduce some of the perceptual biases when visually validating a trend line, there is no single solution that can remedy all the problems. 
%These results together challenge the benefits of current practice of visual model validation as a part of the data analysis process and suggest the need for further research on this topic.

The paper is structured as follows: After an overview of current research studies on visual validation and visual estimation in \cref{sec:rw}, we give details on the general data, study, and analysis design in \cref{sec:study}. The two experiments - validation versus estimation and visual designs- as well as their analysis and results are presented in \cref{sec:exp1} and \cref{sec:exp2}. \cref{sec:futurework} discusses limitations of our work and possible future work. 

%% file: text/sec-rw.tex
\section{Related Work}
\label{sec:rw}

\paragraph{Visual Model Validation}

The research topic of visual model validation has received little attention in the past, but has become more prominent recently as it is recognized as an essential part of exploratory data analysis~\cite{Hullman.2021}. Braun et al.~\cite{Braun.2023} were the first to investigate the perceptual differences between visual validation and visual estimation using the example of average values in scatterplots. They found that participants were more accurate in estimation than in validation, that the visual validation of averages is unbiased, and that the critical point between accepting and rejecting a given value is close to the statistical 95\% confidence interval. This study motivated this work on the validation of linear trends.

Few other studies relate to our experiment. Majumder et al.~\cite{Majumder.2013} examine the visual validation of statistical inference in linear models, in which participants had to visually find the most deviating model (e.g. highest slope) in a small multiple setting. Their study showed that visual tests have higher power than conventional tests when the effect size is large. The findings of Correll et al.~\cite{Correll.2018} on visual validation of data distributions in scatterplots, histograms, and density plots suggest problems with overplotting, which informed the stimuli design in our study. 

Most of the time, the possibility for visual model validation is given in an interactive way as part of a visual analytics or machine learning system. Chatzimparmpas et al.~\cite{Chatzimparmpas.2020} give an overview of the current use of visualization for machine learning model interpretation. B{\"o}gl et al.~\cite{Boegl.2013} provide a visual analytics process to assist domain experts in selecting suitable models in time-series analysis. While the Visual (dis)Confirmation tool by Choi et al.~\cite{Choi.2019} allows users to perform data analysis by automatically generating appropriate visualizations based on hypotheses framed in natural language, Chegini et al.~\cite{Chegini.2018} support users in identifying local patterns in large scatterplot spaces by automatically comparing local regions using a model-based pattern descriptor. M{\"u}hlbacher and Piringer~\cite{Muehlbacher.2013} introduce a partition-based framework for validating regression models both qualitatively via visualizations and quantitatively via a relevance measure for ranking features.
Our work is synergistic with the prior work in that it contributes to a better understanding of visual model validation that can improve the development of visual analytics and machine learning systems in the future.

An additional component of model validation is the viewer's trust in these models and their visualizations, which has been widely studied in the past~\cite{Rogowitz.1996, Peskov.2020, Mayr.2019, Elhamdadi.2022, Dasgupta.2017}. The survey by Chatzimparmpas et al.~\cite{Chatzimparmpas.2020a} summarizes the importance of this relationship between visual design and trust in machine learning. In our study, we tried to minimize the influence of trust by giving the participants as less information as possible without impairing their understanding of the tasks. 

\paragraph{Visual Model Estimation}

Most studies on the perception of statistical models aimed to understand visual estimation. In addition to research on the estimation of average values~\cite{Hong.2022, Gleicher.2013, Yuan.2019}, there are many papers on the estimation of linear trends.

Correll and Heer~\cite{Correll.2017} examined the basic perceptual process of visual estimation and had participants perform ``regression by eye'' for linear trends in scatterplots and other visualizations. They find that an individual's ability to estimate the slope of a linear trend with respect to the least squares regression model depends on both visual features and data features, without bias for positive and negative trends. Ciccione and Dehaene~\cite{Ciccione.2021} were also interested in the accuracy and bias of visual regression estimation in scatterplots. Their results indicate that people consistently overestimate trends. These works played a key role in our study, hypotheses, and stimuli design.

A different aspect of linear trends is the correlation of the two data dimensions. Rensink and Baldridge~\cite{Rensink.2010} investigated the influence of statistical properties on the perception of correlation in scatterplots. Besides statistical properties, correlation perception research dealt with the influence of visual designs, features, and ensemble coding~\cite{Szafir.2016}. Yang et al.~\cite{Yang.2019} showed that visual features, such as bounding boxes, are used as proxies for estimating correlation in scatterplots. Xiong et al.~\cite{Xiong.2022} found that people estimate correlations more accurately in scatterplots with generic axis labels than with semantic labels. Comparisons of different correlation visualizations based on Weber's law ranked scatterplot as the best visualization design and showed that performances of correlation estimations differ for positive and negative correlations~\cite{Harrison.2014, Kay.2016}. These studies confirm our choice of scatterplots as the visual design for our studies, and their results have implications for our stimuli design.  

%% file: text/sec-study.tex
\begin{figure*}%[tbh!]
    \centering
         \begin{subfigure}[t]{0.22\textwidth}
             \centering
             \includegraphics[width=\textwidth]{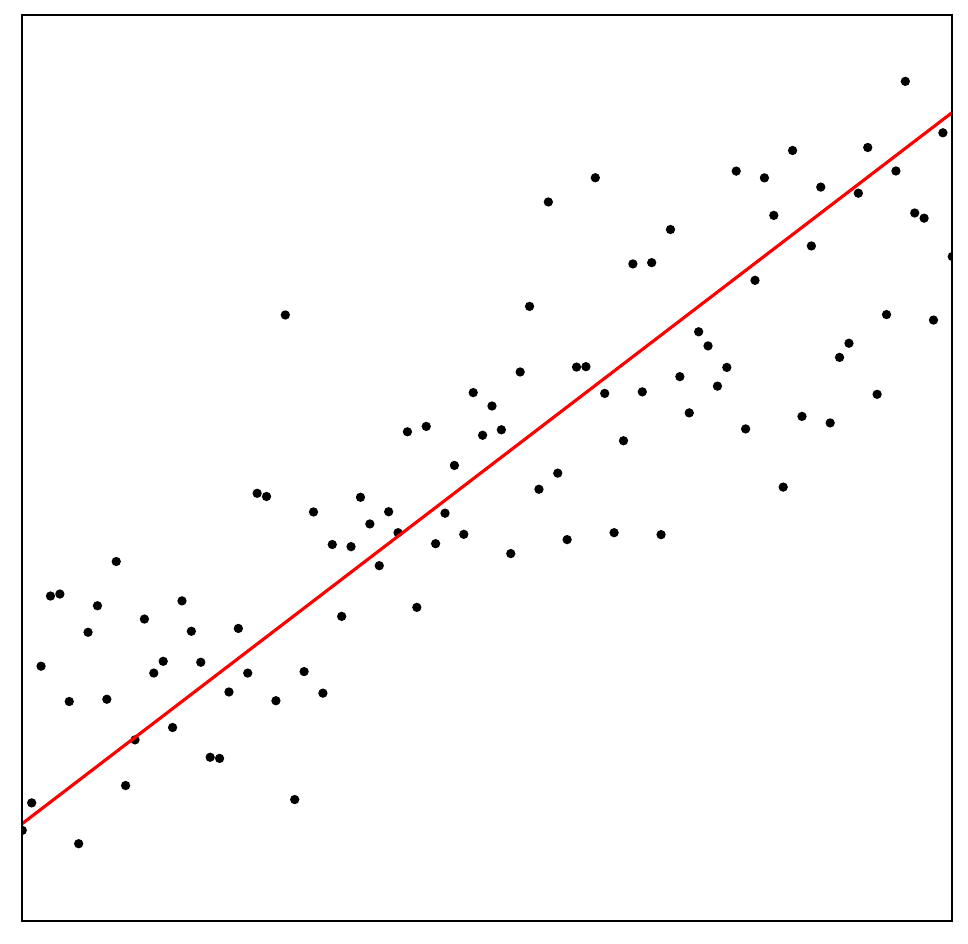}
             \caption{Positive slope deviation of $0.5$.\\~}
             \label{fig:val_dev_pos}
         \end{subfigure}
         %\hfill
         \hspace{1cm}
         \begin{subfigure}[t]{0.22\textwidth}
             \centering
             \includegraphics[width=\textwidth]{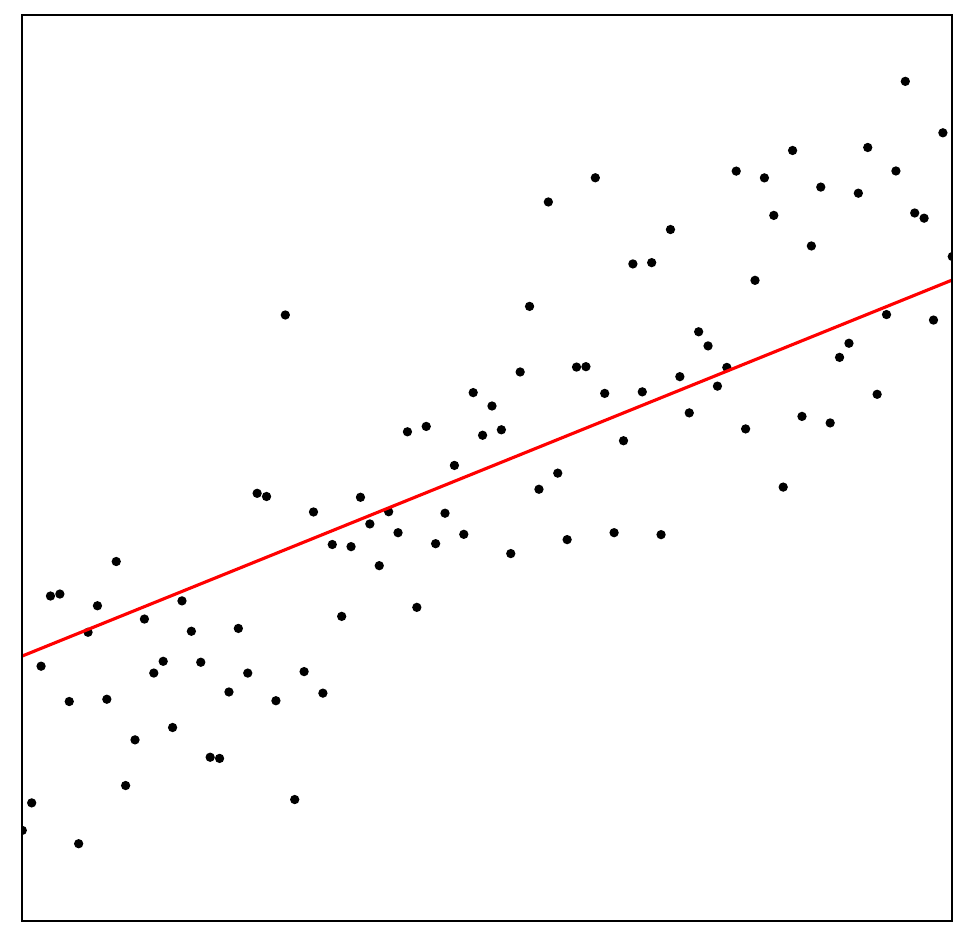}
             \caption{Negative slope deviation of $0.5$.\\~}
             \label{fig:val_dev_neg}
         \end{subfigure}
         %\hfill
         \hspace{1cm}
         \begin{subfigure}[t]{0.22\textwidth}
             \centering
             \includegraphics[width=\textwidth]{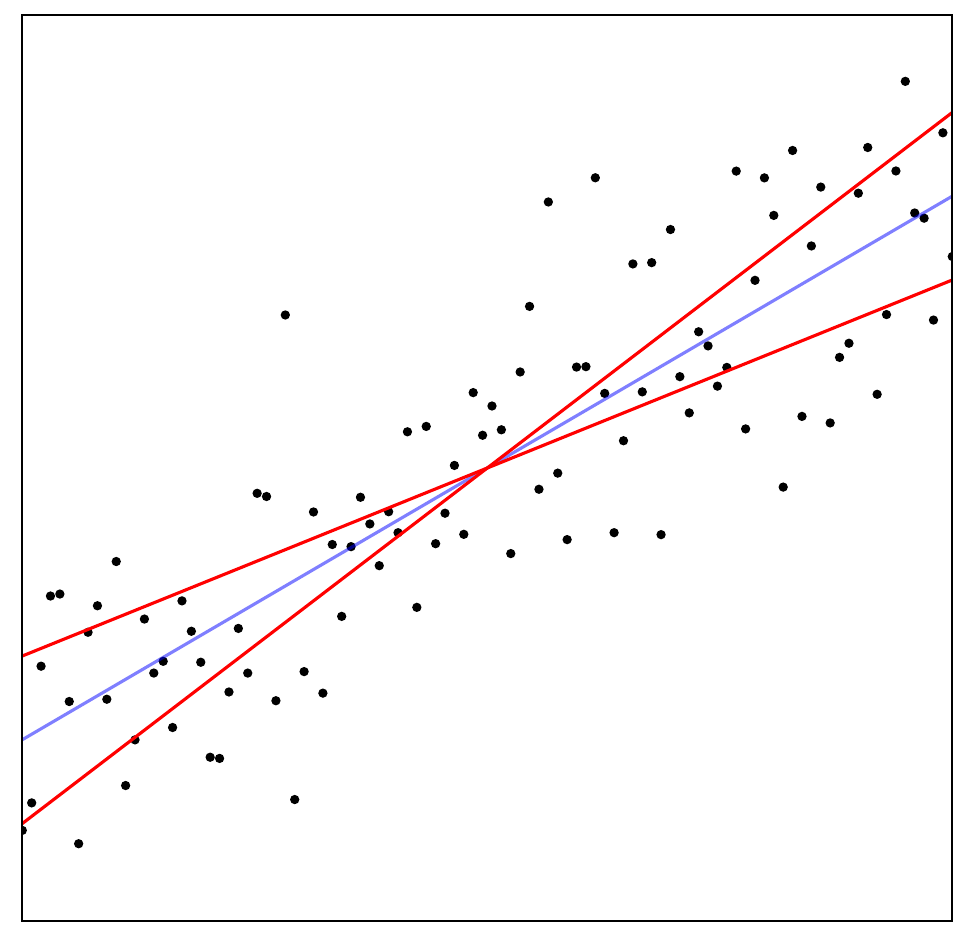}
             \caption{True regression line (blue) and the two deviating lines from (a) and (b) (red).}
             \label{fig:val_dev_corr}
         \end{subfigure}
            \subfigsCaption{Example of the same amount of positive and negative slope deviation in the validation task.}
            \label{fig:val_dev}
\end{figure*}

\section{Experimental Design}
\label{sec:study}

We conducted two experiments to gain insights into the perceptual process of the visual validation of linear trends and to answer our research questions:

\smallskip
\noindent \textbf{RQ1}: How does performance in visual validation of linear trends relate to the accuracy of visual estimation? and

\smallskip\noindent
\textbf{RQ2}: Can common visual designs enhance the performance of visual validation of linear trends? 
\smallskip

The \textbf{first experiment} aimed to answer RQ1. Therefore, it contained two \textit{tasks} to compare the following perceptional processes:
\begin{itemize}[noitemsep, leftmargin=*]
    \item \textit{Visual validation}: 
    Participants were shown scatterplots with an already drawn trend line (see \cref{fig:teaser_val}). They were asked to indicate whether the line was ``too steep',' ``too flat'', or ``about the same'' (i.e., the shown line represents the actual trend) in relation to the true slope of the linear trend of the data.
    \item \textit{Visual estimation}: Participants were asked to fit trend lines to the data in the scatterplots (see \cref{fig:teaser_raw}) by adjusting the slope of the line by moving a slider.
\end{itemize}

The \textbf{second experiment} aimed to answer RQ2 by evaluating three regression visualization designs for validation. In this study, participants only had to perform the validation task described in experiment~1.

\subsection{Study procedure} 

The two experiments addressed different research questions, but used the same study structure and data.

The \textit{study procedure} began with demographic questions about the participants, followed by a short training period for the participants to familiarize themselves with the study interface. To avoid bias in the participants' responses, we did not provide training feedback. To minimize learning effects, the order of trials was randomized. In the study interface, each page displayed one trial (i.e., on plot). For further analysis, response times were recorded. At the end, we asked the participants for their strategy in performing the task and to rate the difficulty of the tasks on a 5-point Likert scale (1 -- very difficult, 2 -- difficult, 3 -- neutral, 4 -- easy, 5 -- very easy)~\cite{Vagias.2006}.

Using a \textit{between-subject study design} (i.e. the participants were divided into two groups and had to either estimate or validate trends), we prevented learning effects between the trials and reduced the number of trials per participant~\cite{Charness.2012}. To ensure consistency and comparability, the same data were used in both studies for all of the between-subject groups. The only difference between the two experiments was the visual representation of the data in the scatterplots (see \cref{fig:teaser_val} and \cref{fig:teaser_error} as an example).

\subsection{Data Generation and Stimuli Design}
\label{sec:data}

Our data generation is inspired by the approaches used by Braun et al.~\cite{Braun.2023} and Correll and Heer~\cite{Correll.2017}. Each trial displayed a scatterplot with a size of $700 \times 700$ pixels. Therefore, we recommended a screen size of 13'' or larger. The scatterplots contained 100 data points in the range of $[0, 1] x [0, 1]$ uniformly distributed along the x-axis. 
We used the standard regression model for the point generation in order to be able to compare the validation and estimation results with the ordinary least squares (OLS) regression: $y=ax+b$. We used this function to generate set of points along particular trends. The trend lines were centered in the scatterplot (as by Correll and Heer~\cite{Correll.2017}), i.e., $f(0.5)=0.5$, and both positive and negative slopes were used. The y-coordinates of the resulting data points were then permuted using a normal distribution. Investigating the perception of the slope of linear regressions, the slope parameter \textit{a} was the only variable in our studies. Centering the target trends in the plot ensured that participants could always estimate the true regression line by solely manipulating the slope value. 

To keep the difficulty level as similar as possible between trials, the standard deviation of the normal distribution was fixed to $0.1$ and the slopes of the trends were set to a range of $[0.35, 0.65]$. Since the permutation of the y-coordinates could lead to a deviation of the resulting regression from the original target trend, we performed rejection sampling to ensure that the slope of the trend of the resulting points was within $10^{-3}$ of the target slope.

To be able to measure the accuracy of visual validation, we showed the participants trend lines with slopes deviating from the true regression slope. Due to the centering of the underlying trends, a variation of the slope means a rotation of the line around the point $(0.5, 0.5)$. We define the \textbf{deviation} of the displayed line as the deviation from the true slope as a proportion of the regression's standard error:
\begin{equation}
    \mbox{shown slope} = \mbox{true slope} + \mbox{\textit{deviation}} \cdot \mbox{standard error}
    \label{eq:dev}
\end{equation}

\cref{fig:val_dev} shows an example validation trial with the same amount of positive and negative slope deviation ($0.5$ in this example).
The standard error of the regression is data dependent and represents the average distance between the points and the regression line. This allows us to compare and analyze trials with different data sets regardless of their properties, since an absolute change in slope is perceived differently in graphs with high and low point dispersion.

The deviation definition is based on the calculation of the regression confidence interval \textit{(CI)}. In statistical analysis, the 95\% CI is a common measure for the uncertainty in models~\cite{Altman.2000}. By construction -- in our setting with a fixed number of points and distribution -- the slope's 95\% CI is set at a deviation of $0.198$. We use the confidence interval to compare the user study results with statistical quality measures. In a statistical sense, all lines with a smaller slope deviation should be considered acceptable. However, people may have a smaller confidence interval and reject lines with deviations less than 0.198.

Based on the result of the study for validation and estimation of the average value~\cite{Braun.2023}, where deviations greater than $0.7$ were consistently rejected, we used the same deviation range of $[-0.7,0.7]$ (i.e., we showed lines with a maximum slope deviation of $0.7$ based on \cref{eq:dev}). The deviations we used for the lines shown in the studies were evenly distributed within this range in $0.05$ increments (including deviation $0$), resulting in a total of \textbf{58 trials} (i.e., 58 different data sets) per participant (29 trends with a positive slope, 29 with a negative slope).

In line with the literature~\cite{Braun.2023, Correll.2017, Xiong.2022}, we kept the displayed scatterplots as clean as possible to minimize visual distraction by omitting axes marks and labels (see example stimuli in \cref{fig:teaser}). The used colors and marker sizes were the same as in the study by Braun et al.~\cite{Braun.2023}.

\subsection{Analysis Procedure}

In order to assess our results in relation to literature, our analysis is similar to that of Braun et al.~\cite{Braun.2023}. 
For comparability of the estimation and validation results, we transformed a participant's validation responses to binary results: \textit{1} for \textit{accepting} (i.e., "about the same") the shown line, \textit{0} for \textit{rejecting} (i.e., "too steep" or "too flat")  the shown line. Similar to previous studies, logistic regression was then applied to the acceptance rates~\cite{Yang.2019, Braun.2023} to assess validation accuracy.

The estimation errors (i.e., the deviation in slope of the self-adjusted trend lines) were calculated using the same deviation definition as for validation (\cref{eq:dev}). This allows us to compare the logistic regression of validation acceptance rates with the cumulative distribution (CDF) of estimation errors.

For statistical testing, we used a multi-stage approach with the standard significant level $\alpha=0.05$ in all tests. First, we performed a Shapiro-Wilk test on the given responses and response times to test the data for normality, with the result that none of the data fulfilled this property. Based on this, we then used the non-parametric Kolmogorov-Smirnov \textit{(KS)} test for the difference in validation and estimation results (i.e., comparing the logistic regressions of the validation acceptance rates and the CDF of the estimation errors). A Wilcoxon test was uses to test the response times and estimation errors on difference in means. For comparing the Likert responses, we used chi-squared test. For experiment~2 we first performed a Kruskal-Wallis test on the response times and a chi-squared test on the categorical responses to test the visual designs for significant differences. As post-hoc pairwise analysis of the respective tests to compare the different design combinations.

%% file: text/sec-exp1.tex
\section{Experiment 1: Visual Validation versus Visual \\Estimation}
\label{sec:exp1}

Experiment 1 analyzed the performance in visual validation of linear trends in relation to visual estimation. Based on previous research on model perception~\cite{Braun.2023, Proffitt.1995, Correll.2017, Ciccione.2021} and our own assessments when generating the data, we propose the following hypotheses for experiment~1:

\begin{itemize}[itemsep=0pt, parsep=3pt, leftmargin=*]
    \item \textbf{H1}: \textit{The accuracy of visual validation is lower than the accuracy of visual estimation when perceiving the slope of a linear trend in a scatterplot.} \\
    The study by Braun et al.~\cite{Braun.2023} for the average value showed that visual estimation provides a higher level of accuracy compared to visual validation. We expect this result to hold for linear trends.
    \item \textbf{H2}: \textit{People's critical point between accepting and rejecting a given trend line when validating is close to the boundary of the 95\% CI.} \\
    Also based on the results of Braun et al.~\cite{Braun.2023} we expect people's ``visual confidence'' to match the statistical CI, as is true for the visual validation of average values.
    \item \textbf{H3}: \textit{For visual validation, the results don't differ between positive and negative trends.}\\
    Correll and Heer~\cite{Correll.2017} found no bias in visual estimation of linear trends. We expect visual validation and estimation to be similar in this regard.
    \item \textbf{H4}: \textit{For visual estimation, people overestimate the slope of linear trends.} \\
    We expect this behavior found in previous research~\cite{Proffitt.1995, Ciccione.2021, Correll.2017} to be the same in our study.
    \item \textbf{H5}: \textit{Perceived task difficulty and task completion time are lower for visual validation than for visual estimation.} \\
    The validation task is simply a matter of acceptance, whereas the estimation task requires participants to fit a line to the data. Therefore, we expect participants to perceive the validation task as easier and complete it faster.
\end{itemize}

\subsection{Experimental Setting and Participants}

We conducted an online study on Limesurvey~\cite{LimeSurvey.2024} and recruited participants from the crowdsourcing platform Prolific~\cite{Prolific}. A total of 122 participants took part in the study. They had to speak English fluently. No restrictions were made for country of residence. We removed the data from 12 participants because of their incorrect answers to the attention question. Out of the 110 remaining participants, 46 completed the validation and 64 answered the estimation task. Most of them were between 20 and 40 years old (86\%) and the gender distribution was close to even (F: 48\%, M: 49\%, other: 3\%). Participants' educational levels ranged from high school diplomas to doctorate degrees, with the majority having a bachelor's degree (40\%). The overall self-reported expertise in statistical model estimation was relatively low, with 88\% of the participants indicating an expertise between 1 and 3 on a 5-point Likert scale.
The average time to complete the study was 20 minutes. Participants were compensated with \pounds4.45.

\subsection{Results}

Given our data generation approach based on the ordinary least squares (OLS) regression (see \cref{sec:data}), we analyze the responses of the study participants based on this model.

\subsubsection{Accuracy of Visual Validation vs. Visual Estimation}

We evaluate the accuracy of visual validation and estimation in relation to the ``statistical accuracy'' of OLS regression. In statistical terms, all trend lines within the 95\% confidence interval are considered to be valid. This means that, in a perfect world, participants should accept all trend lines with a slope deviation of less than 0.198 (i.e., the CI slope deviation) and reject all trend lines with a higher slope deviation in the validation task. In the estimation task, participants should ideally only estimate lines with slope deviations less than 0.198.

\begin{figure}%[tbh!]
 \centering
 \includegraphics[width=0.7\columnwidth]{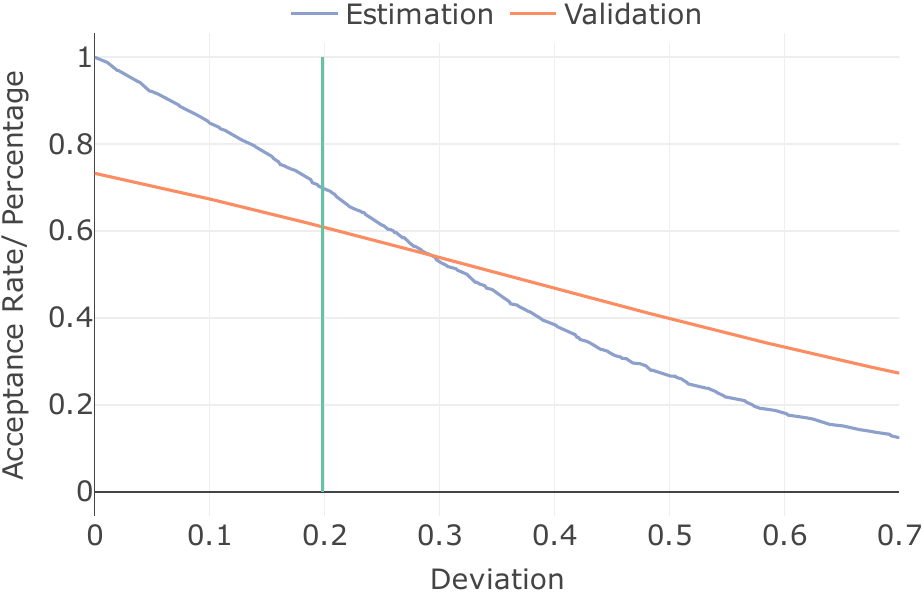}
 \caption{Comparison of validation and estimation accuracy (absolute deviation) with respect to OLS regression. Blue line: Cumulative distribution (CDF) for the estimation errors. Orange line: Logistic regression for the validation acceptance. Green line: Statistical 95\% CI. Notice that more statistically valid lines were estimated than accepted by validation and more invalid lines were accepted than estimated.}
 \label{fig:val_est_comp}
\end{figure}

To be able to compare validation and estimation results, we summarize the acceptance rates and estimation errors. We combine the results for positive and negative trends and positive and negative slope deviations as absolute acceptance rates and errors. \cref{fig:val_est_comp} shows the resulting logistic regression for the validation acceptance rates and the cumulative distribution for the estimation errors.

As shown, when the slope deviation is low (e.g., less than the 0.198 -- the green line), participants were more accurate in estimating than validating trend lines.
For example, about 70\% of the participants were able to estimate (i.e., drew) trend lines with slope values below the deviation of 0.198. 
In contrast, only 60\% of the participants correctly validated (i.e., accepted) shown trend lines with the same slope values.
When the slope deviation is high, again we see that participants were more accurate in estimating than validating trend lines.
For example, only about 20\% of the participants estimated (i.e., drew) trend lines with slope values above the deviation of 0.6.
In contrast, about 35\% of the participants falsely validated (i.e., accepted) shown trend lines with the same slope values.
Only in the deviation range adjacent to the confidence interval ($0.2$ to $0.3$) validation is more accurate than estimation. In this statistically invalid range, a slightly higher percentage of lines were estimated than accepted.
Notably, the validation acceptance of linear trends has an almost linear relationship with the slope deviation. Ideally, it should have high values at low slope deviations with sharp drop at the 95\% CI border (i.e., $0.198$). 

The KS-test showed the two curves to be significantly different ($p\ll0.01$, $D=0.296$). Therefore, \textbf{H1} is supported for OLS regression. 

%The difference in the curves is clearly visible. The proportion of trend lines drawn with small slope deviation is higher than the acceptance rate of the validated lines with the same slope deviation. Similarly, the acceptance rate of the validated lines with a large slope deviation is higher than the proportion of trend lines drawn. This indicates that participants were more accurate estimating than validating. 

The critical points of validation and estimation are precisely these slope deviations, with a 50/50 chance that a line will be accepted or rejected, or estimated with a lower or higher slope. This critical point is slightly lower for estimation ($crit_{val} \approx 0.36 > crit_{est} \approx 0.32$). Moreover, the critical point of validation is much greater than the deviation of the 95\% CI ($0.198$), which does not support our hypothesis \textbf{H2} for the OLS regression.
The critical points describe an experimental human threshold for the two tasks. People estimated lines that would be correct with the statistical 99.8\% CI. For visual validation, they even accepted lines with a deviation greater than the statistical 99.9\% CI.

When analyzing the individual performance of the participants in visual validation, it is noticeable that 72\% of the participants had an individual critical point greater than the 95\% CI and 13\% accepted incorrect models more often than correct ones.

\subsubsection{Bias in Positive and Negative Slope and Deviation}

\begin{figure}%[tbh!]
    \centering
    \begin{subfigure}[t]{0.7\columnwidth}
         \centering
         \includegraphics[width=\columnwidth]{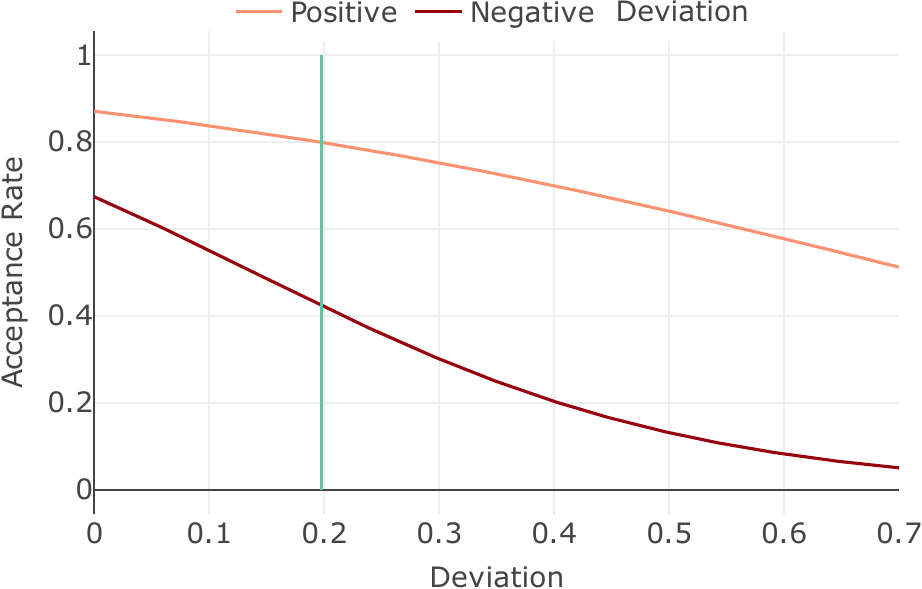}
         \caption{Positive slope.}
         \label{fig:val_error_comp_positive}
     \end{subfigure}
    
     \begin{subfigure}[t]{0.7\columnwidth}
         \centering
         \includegraphics[width=\columnwidth]{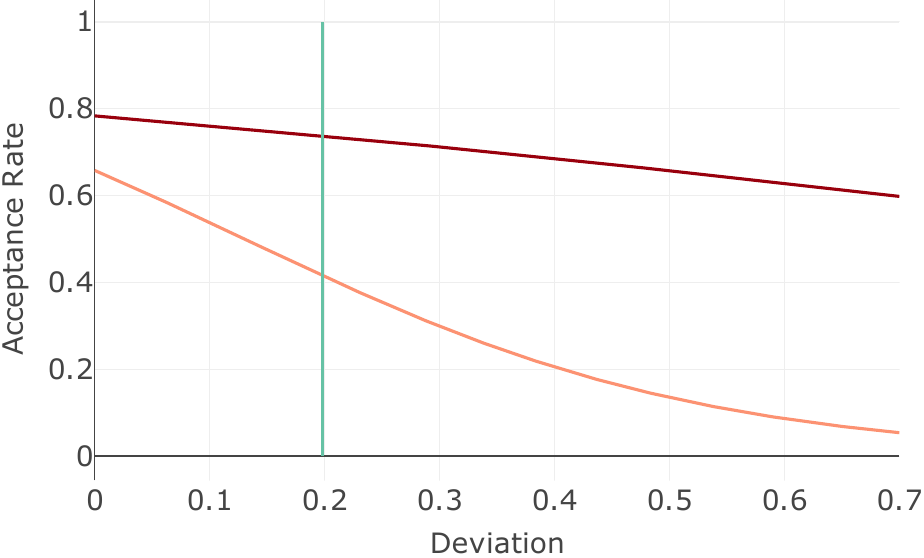}
         \caption{Negative slope.}
         \label{fig:val_error_comp_negative}
     \end{subfigure}
        \caption{Comparison of the \textit{validation} acceptance rates of positive and negative deviations for positive and negative trends with respect to OLS regression. Green line: statistical 95\% CI.}
        \label{fig:val_error_comp}
\end{figure}

\cref{fig:val_error_comp} differentiates the validation acceptance rates by positive and negative trends as well as positive and negative slope deviation. For positive trends, lines with positive slope deviations (i.e., trend lines that were ``too steep'') were accepted significantly more often than lines with negative deviations (see \cref{fig:val_error_comp_positive}) (KS: $p\ll0.01$, $D=0.814$). For negative trends, the results are mirrored, i.e., lines with a negative slope deviation were accepted significantly more often (\cref{fig:val_error_comp_negative}) (KS: $p\ll0.01$, $D=0.926$). For both trend directions, lines that were ``too steep'' were still accepted more than 50\% of the time, even with the largest slope deviation. 
Analysis of individual participant acceptance rates showed that the difference between positive and negative slope deviations was significant for each individual participant.
Comparing the logistic regressions of the acceptance rates between positive and negative trends, there is no significant difference for ``too flat'' lines (KS: $p>0.99$, $D=0.022$), but a significant difference for ``too steep'' lines (KS: $p\ll0.01$, $D=0.336$), indicating a slightly more accurate validation of positive trends. Thus, hypothesis \textbf{H3} cannot be rejected.

\begin{figure}%[tbh!]
 \centering
 \includegraphics[width=0.7\columnwidth]{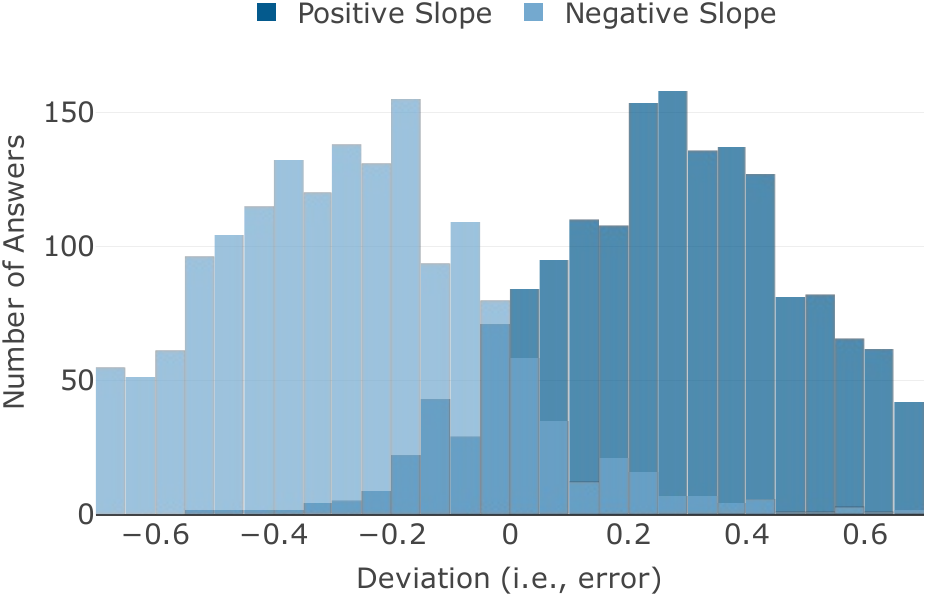}
 \caption{Histogram of the deviations of the \textit{estimated} lines for positive and negative trends with respect to OLS regression.}
 \label{fig:est_error}
\end{figure}

A similar pattern can be observed for the estimation errors (see \cref{fig:est_error}), supporting the results of previous studies~\cite{Ciccione.2021, Proffitt.1995, Correll.2017} and our hypothesis \textbf{H4}: People overestimate the slope of trend lines. Without over-estimation, both distributions should be centered at 0 slope deviation. With an average slope deviation of $0.357$ for positive trends and $-0.359$ for negative trends, participants consistently drew the trend lines too steeply (Wilcoxon: $p_{\mbox{positive}}\ll0.01$, $V_{\mbox{positive}}=1660685$; $p_{\mbox{negative}}\ll0.01$, $V_{\mbox{negative}}=78910$). No significant differences in estimation errors between positive and negative trends could be found (Wilcoxon: $p>0.52$, $W=1743165$).

In summary, both perceptional processes -- visual validation as well as visual estimation -- are biased toward ``too steep'' slopes for positive as well as negative trends.

\begin{figure}[tbh!]
 \centering
 \includegraphics[width=0.55\columnwidth]{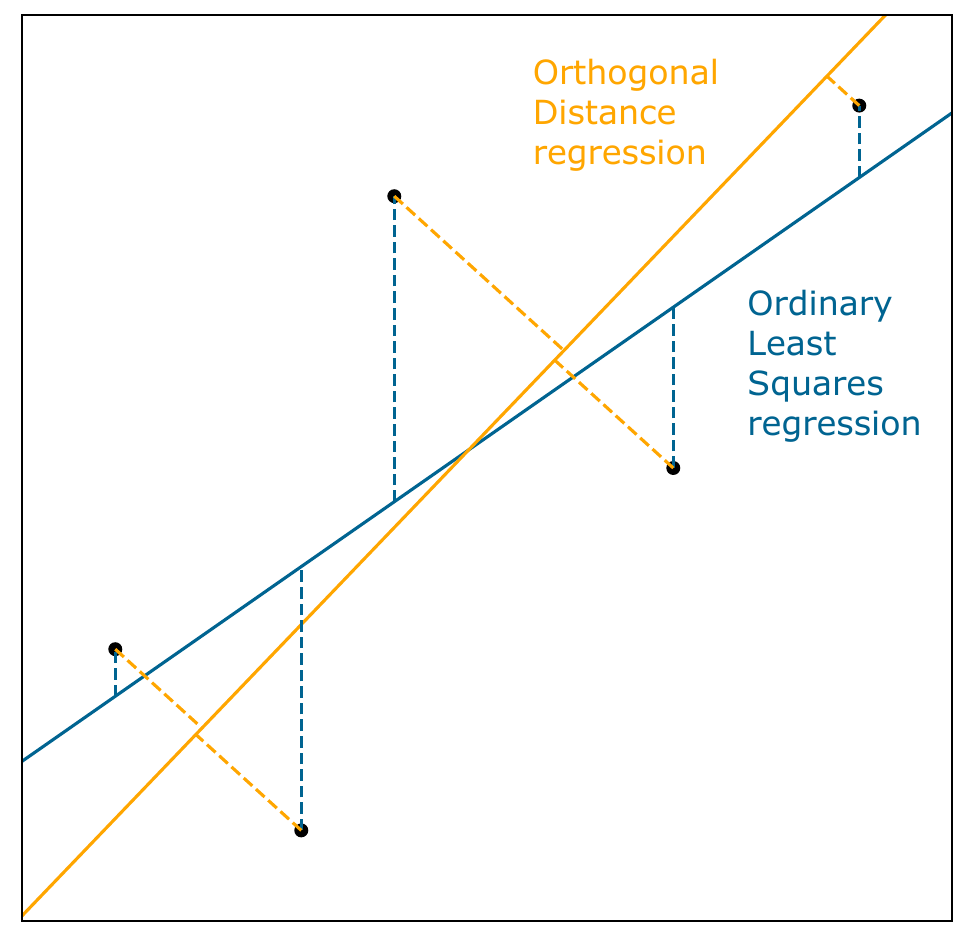}
 \caption{Illustration of the difference between ordinary least squares (OLS, blue line) and orthogonal distance (ODR, yellow line) regression, adapted from~\cite{Ciccione.2021}.}
 \label{fig:ols_odr}
\end{figure}

\subsubsection{Post-Hoc Analysis: OLS vs. ODR}

The results opened a new question: What is the reason why people perform poorly at perceiving trend lines that are ``too steep''?

%The same absolute change in the slope value results in different angle changes of the trend lines for positive and negative slope deviation. Since this difference is only 0.33° on average in our study, this cannot be a reason.
A clue to this question was found in a study by Ciccione and Dehaene~\cite{Ciccione.2021}. Inspired by their results, we hypothesize that individuals perceive orthogonal distance (ODR) instead of orthogonal least squares (OLS) regression, since ODR regression has a higher slope per calculation. \cref{fig:ols_odr} illustrates the difference between the two regression models. OLS regression minimizes the sum of the squares of the vertical distances of the points to the line and assumes noise only in the dependent variable (y-axis). ODR regression minimizes the sum of the squares of the orthogonal distances of the points to the line and assumes noise in both variables.
The slopes of the ODR regression lines differ from those of the OLS regression. Thus, taking the ODR regression model into account changes the true slope value, which influences the measured accuracy of visual validation and estimation. We analyzed and compared the results of both tasks with respect to both regression models -- OLS and ODR -- as a post-hoc analysis to see if it explained the bias in slope perception.

\begin{figure}%[tbh!]
 \centering
 \includegraphics[width=0.7\columnwidth]{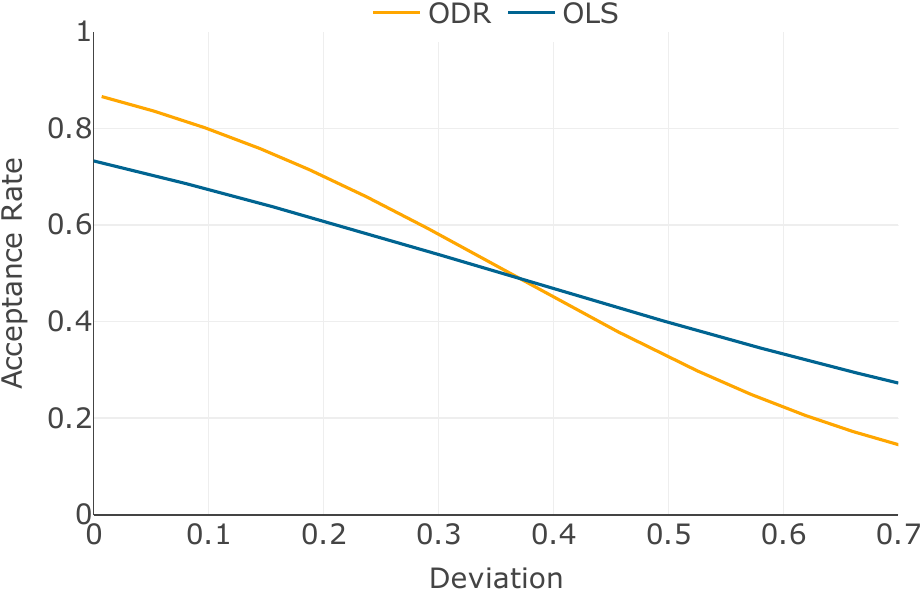}
 \caption{Comparison of \textit{validation} accuracy (absolute deviation) with respect to OLS and ODR regression.}
 \label{fig:regression_comparison_val}
\end{figure}

\smallskip\noindent \textbf{Visual Model Validation:}
The acceptance rates for the shown trend lines in the validation task compared for ODR and OLS are visualized in \cref{fig:regression_comparison_val}. They differ significantly for both models (KS: $p\ll0.01$, $D=0.342$). For ODR regression, more lines with small slope deviation were accepted and more lines with large deviation were rejected. The logistic regressions intersect closely at the 50\% acceptance rate, i.e. the critical point between acceptance and rejection of a line remained almost identical for both regression models.

\begin{figure}
 \centering
 \includegraphics[width=0.7\columnwidth]{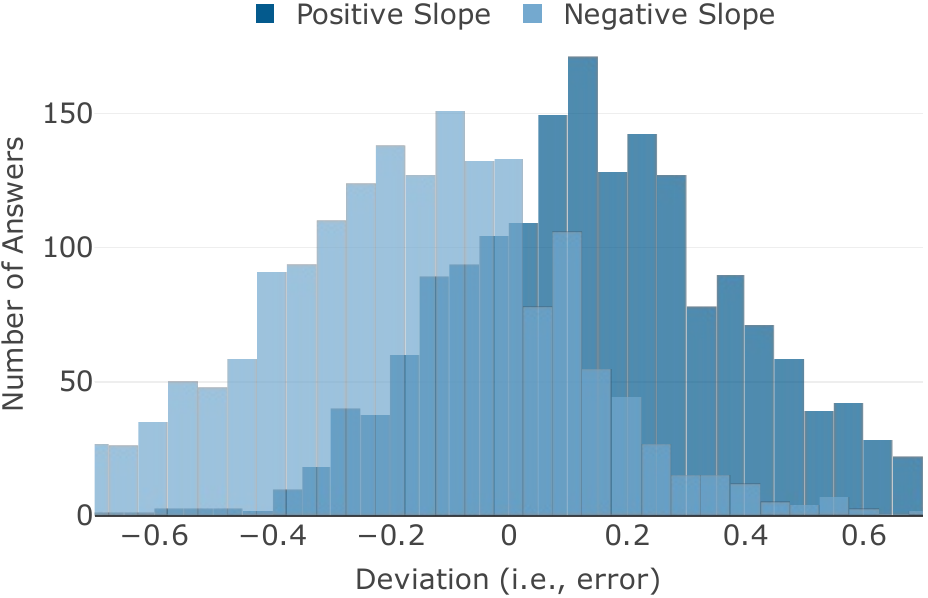}
 \caption{Histogram of the deviations of the \textit{estimated} lines for positive and negative trends with respect to ODR regression.}
 \label{fig:est_error_odr}
\end{figure}

\smallskip\noindent \textbf{Visual Model Estimation:}
There is also evidence of an improvement in visual estimation (see \cref{fig:est_error_odr}). The estimation errors have improved for both positive and negative trends and the slope is less strongly overestimated compared to OLS ($\mu_{pos}=0.210$, $\mu_{neg}=-0.216$).

\begin{figure}
 \centering
 \includegraphics[width=0.7\columnwidth]{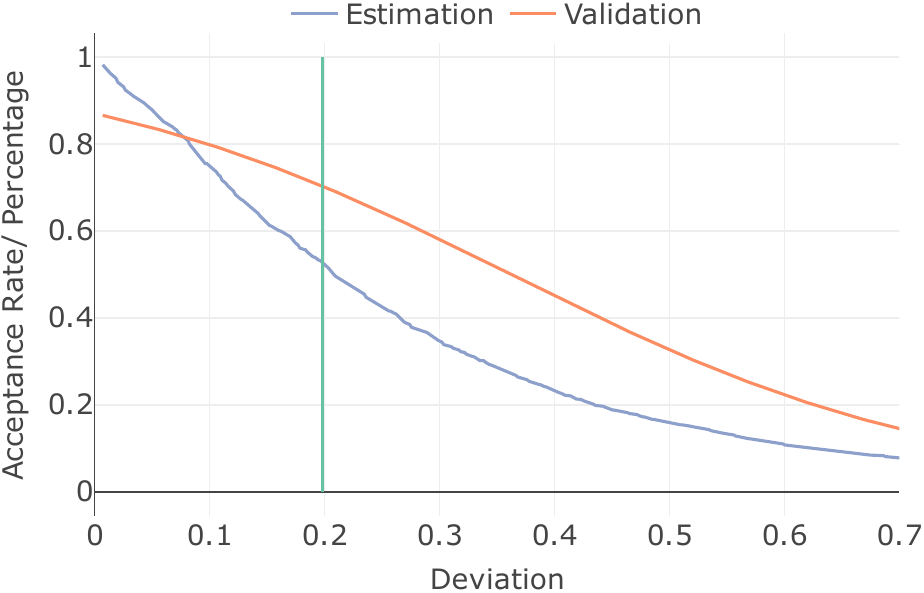}
 \caption{Comparison of validation and estimation accuracy (absolute deviation) with respect to ODR regression. Blue line: Cumulative distribution (CDF) for the estimation errors. Orange line: Logistic regression for the validation acceptance. Green line: Statistical 95\% CI.}
 \label{fig:val_est_comp_odr}
\end{figure}

Comparing the two tasks, the differences between estimation and validation were greater for ODR than for OLS (see \cref{fig:val_est_comp_odr}). The critical point of estimation moved closer to the border of the 95\% CI for ODR ($crit_{est} = 0.208$). As a result, the estimation errors were significantly lower than the acceptance threshold for validation (KS: $p\ll0.01$, $D=0.234$).

\begin{figure}
    \centering
    \begin{subfigure}[t]{0.49\columnwidth}
         \centering
         \includegraphics[width=\columnwidth]{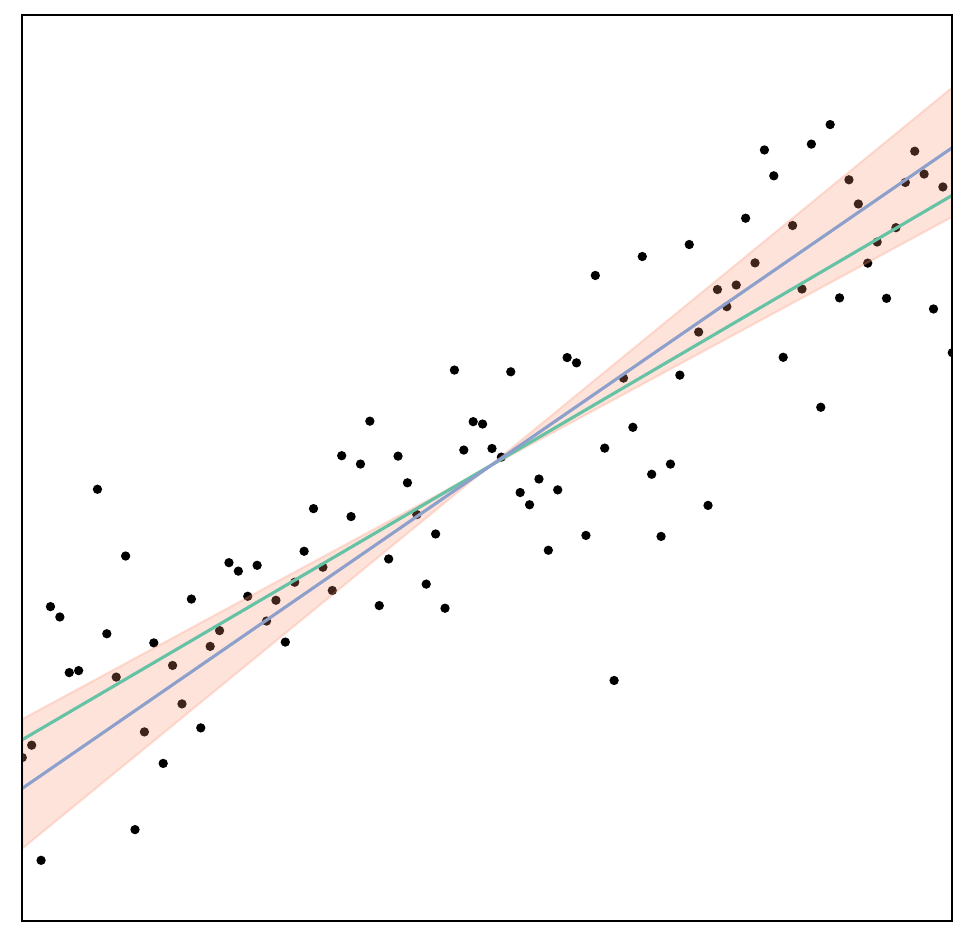}
         \caption{Mean estimation and validation responses with respect to OLS regression.}
         \label{fig:ols_story}
     \end{subfigure}
    \hfill
     \begin{subfigure}[t]{0.49\columnwidth}
         \centering
         \includegraphics[width=\columnwidth]{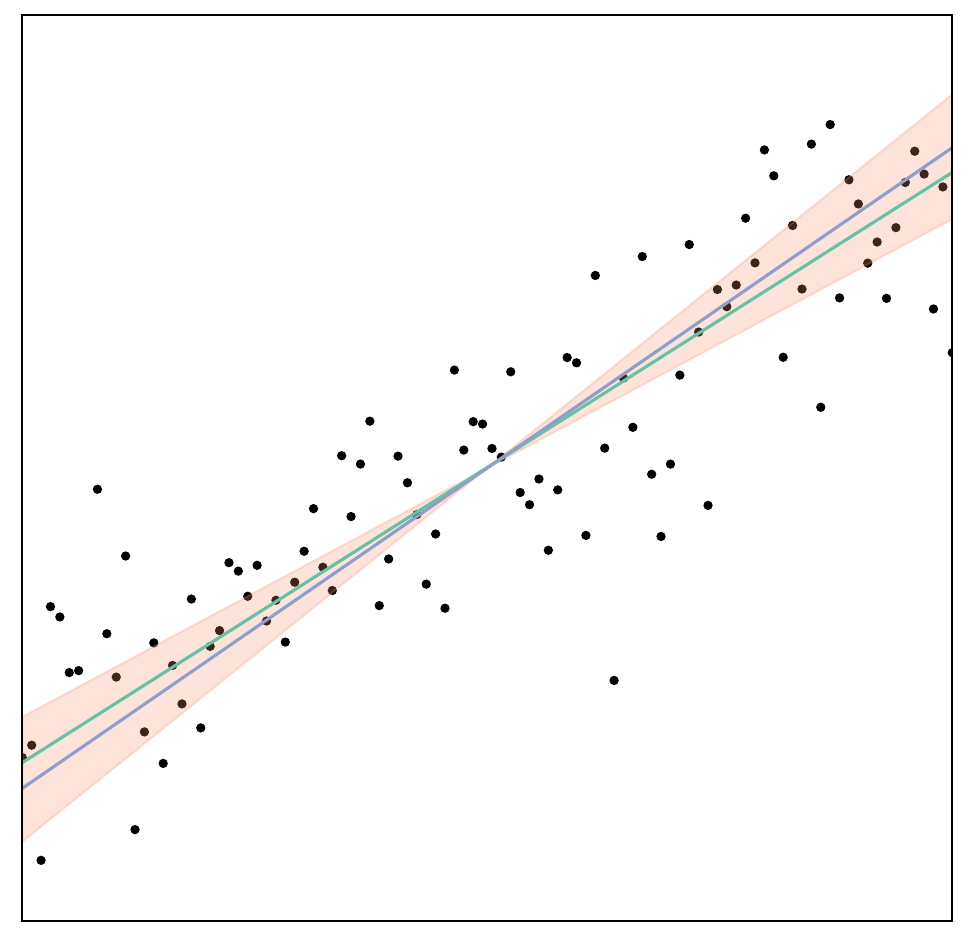}
         \caption{Mean estimation and validation responses with respect to ODR regression.}
         \label{fig:odr_story}
     \end{subfigure}
        \caption{``Visual summary'' of the results for experiment 1 in an example stimulus. The figures show the true regression line (green) for OLS (a) and ODR (b) together with participants' average response for estimation (blue) and the range of lines with an acceptance rate of 50\% or higher for validation (orange). Notice that in figure (b), the blue and the green lines are closer to each other than in figure (a) and the orange range better encapsulate the green line. These suggest that the ODR model better fits participants' perception of trend lines.} 
        \label{fig:results_story}
\end{figure}

\cref{fig:results_story} illustrates the accuracy results of experiment 1. It visualizes an example stimuli with the true OLS and ODR regression lines together with the average responses for validation and estimation. For estimation, this means a line (shown in blue) with the average deviation of the estimated lines. For validation, this is represented as the range of lines (shown in orange) that would be accepted at least 50\% of the time.

In sum, we found that participants' trend perception was more consistent with the ODR regression model than with the OLS model for both estimation and validation tasks.

\subsubsection{Response Time and Difficulty}

\begin{figure}%[tbh!]
    \centering
         \begin{subfigure}[tb]{0.49\columnwidth}
             \centering
             \includegraphics[width=\columnwidth]{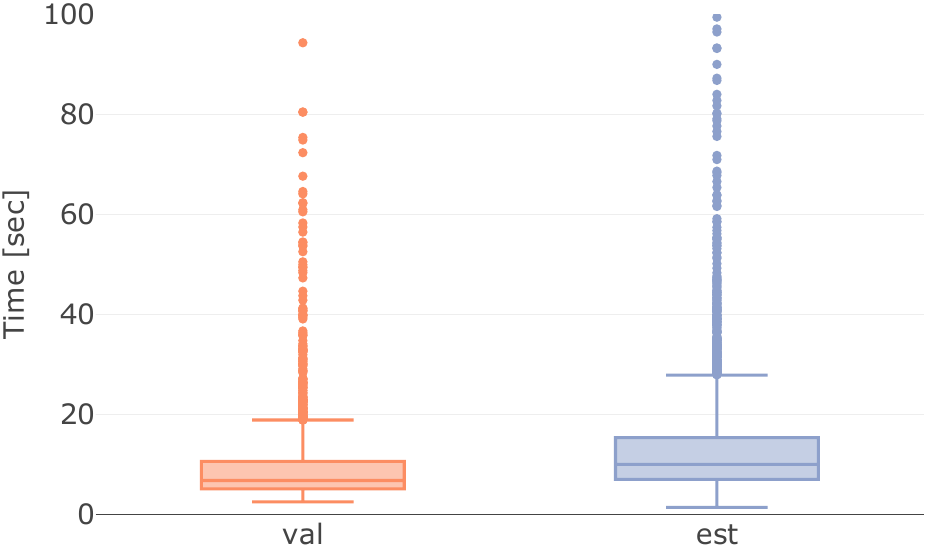}
             \caption{Response times.}
             %\label{fig:exp1_time}
         \end{subfigure}
         \hfill
         \begin{subfigure}[tb]{0.49\columnwidth}
             \centering
             \includegraphics[width=\columnwidth]{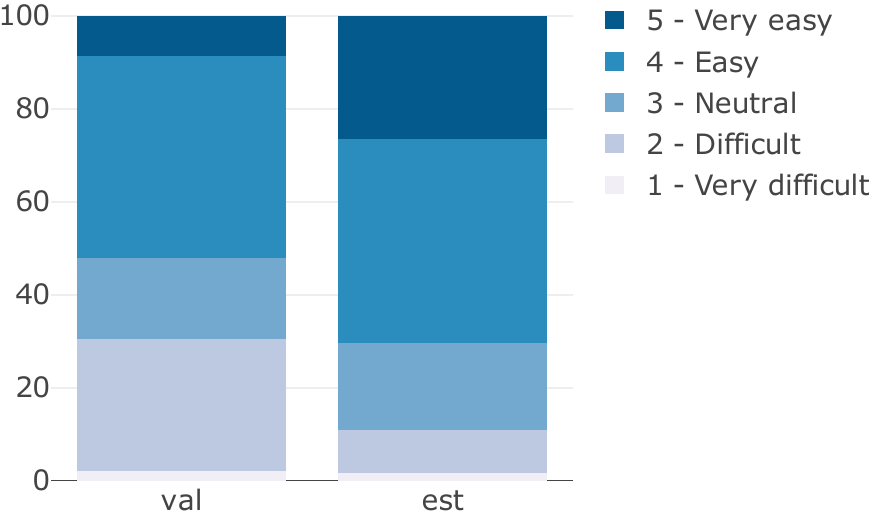}
             \caption{Self-reported task difficulty.}
             %\label{fig:exp1_diff}
         \end{subfigure}
            \subfigsCaption{Response times and self-reported difficulty for experiment 1.}
            \label{fig:exp1_time_diff}
\end{figure}

\cref{fig:exp1_time_diff} shows the response times and the distribution of the Likert scale responses for the task's difficulty. 
As shown, the response times for visual validation were significantly faster than for visual estimation (Wilcoxon: $p\ll0.01$, $\mbox{cohensD}=0.12$, $\mu_{val}=10.85sec$, $\mu_{est}=13.73sec$). In contrast, participants' self-reported task difficulty was significantly lower for the estimation than for the validation task (Chi-squared: $p<0.05$, $\chi^2=10.08$; $\mu_{val}=3.28$, $\mu_{est}=3.84$).
Therefore, Hypothesis \textbf{H5} cannot be rejected.

\begin{figure*}%[tbh!]
    \centering
         \begin{subfigure}[tb]{0.22\textwidth}
             \centering
             \includegraphics[width=\textwidth]{teaser_error}
             \caption{OLS error lines.}
             \label{fig:design_error}
         \end{subfigure}
         %\hfill
         \hspace{1cm}
         \begin{subfigure}[tb]{0.22\textwidth}
             \centering
             \includegraphics[width=\textwidth]{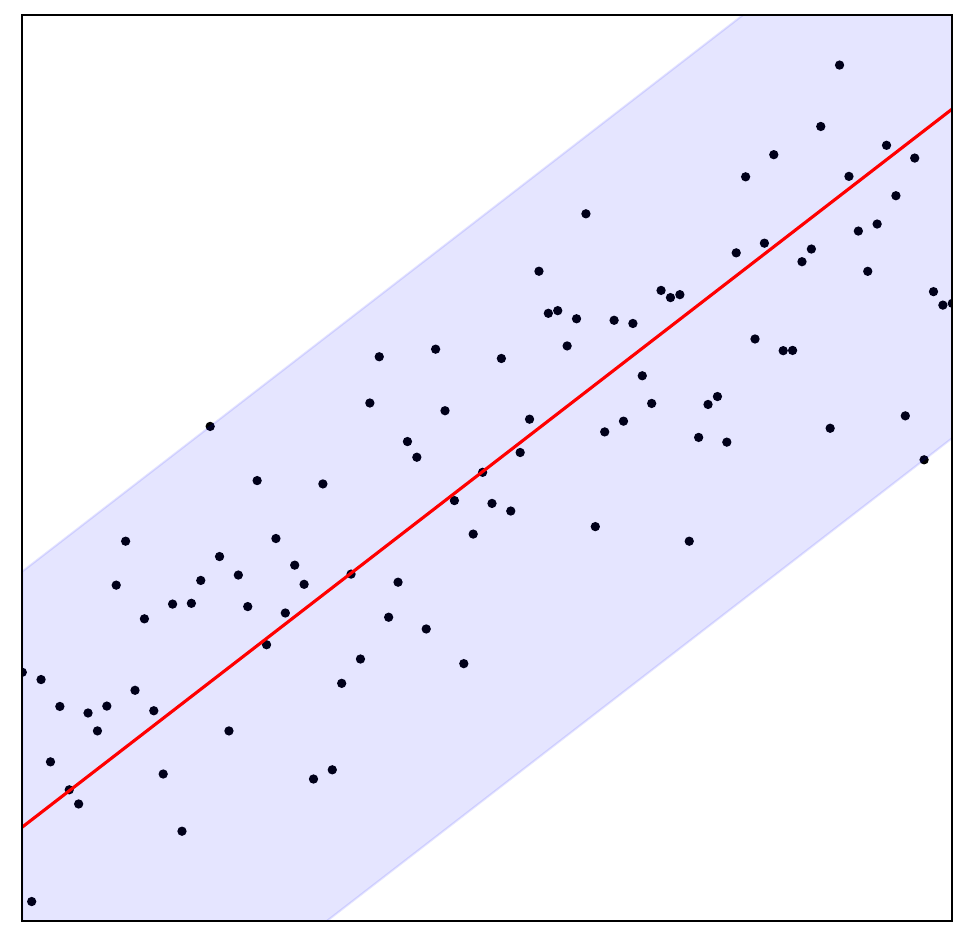}
             \caption{Bounding box.}
             \label{fig:design_box}
         \end{subfigure}
         %\hfill
         \hspace{1cm}
         \begin{subfigure}[tb]{0.22\textwidth}
             \centering
             \includegraphics[width=\textwidth]{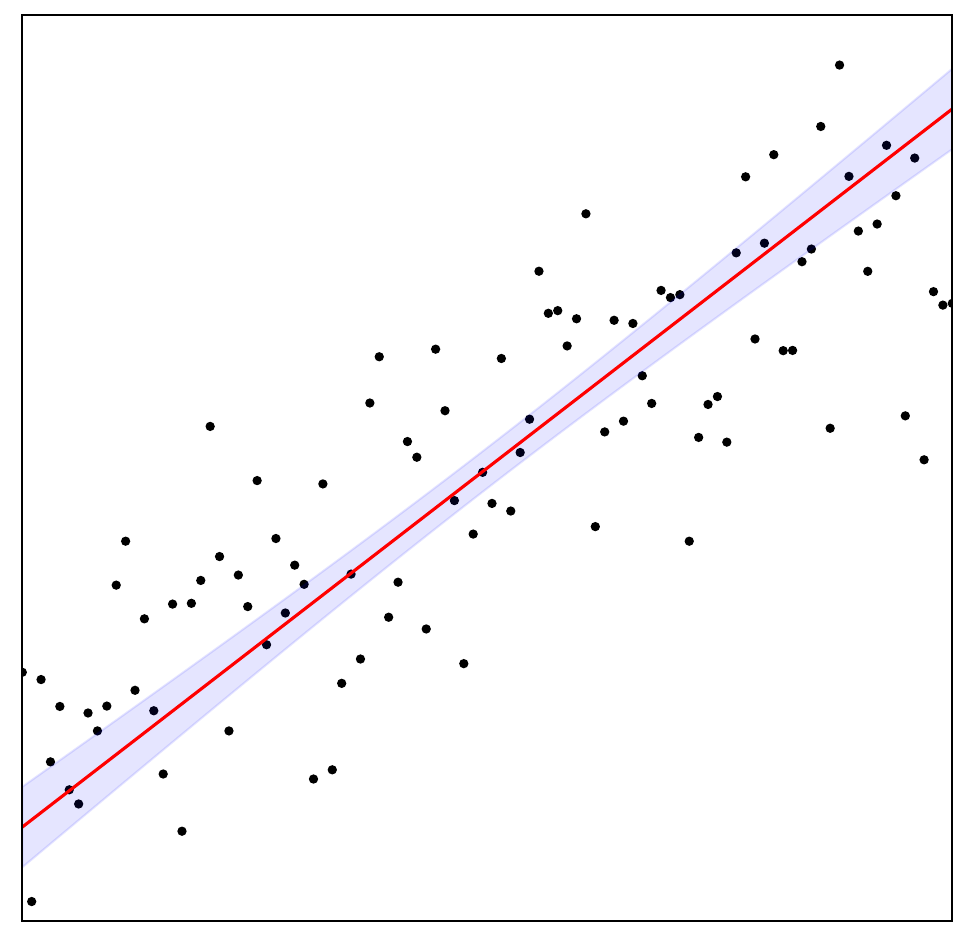}
             \caption{95\% confidence interval.}
             \label{fig:design_conf}
         \end{subfigure}
            \subfigsCaption{Visual designs evaluated for visual validation in experiment 2.}
            \label{fig:designs}
\end{figure*}

\subsubsection{Self-Reported Strategies}
\label{sec:strategies}

Participants' self-reported strategies were provided in free-text form and subsequently summarized by us.
For \textit{validation}, the most often strategy was the comparison of the shown line shown with a self-estimated line (responded by 8 participants). Alternative strategies include a counting strategy ($n=7$) where the participants counted the dots on both sides of the line, and a strategy that references the overall visual image of the visualization ($n=6$) where the participants checked whether the trend line passed through the center of the area of dots. The latter strategy is similar to the strategy of a ``bounding box'' around the scatter dots, which was investigated by Yang et al.~\cite{Yang.2019} for the perception of correlation (see \cref{sec:designs}).

For \textit{estimation}, one strategy was used very often. 14 participants reported using the counting strategy and balancing the number of points on each side of the line. Only two other strategies were mentioned more than once: drawing a line through the middle of the dots ($n=3$) and mentally connecting the dots to a line chart ($n=2$).

For both tasks, the comparison of the results per strategy did not show any significant differences.

%In sum, both computational and visual approaches were chosen for both tasks.

\subsection{Discussion}

Comparing the perceptual tasks of visual validation and estimation, individuals are more accurate at estimating regression models themselves than validating existing ones. One possible reason for this is that, as noted in the self-reported strategies, when validating a trend line, many people compare the line shown to a self-estimated line. Due to their uncertainty in their own estimation, they then may give themselves a margin of tolerance, which is reflected in the larger accepted deviations for the validated lines. This may indicate that people go through a two-step process during visual validation, with estimation as the first step. However, the response times and the self-reported task difficulty provide contradictory supporting evidence. Although participants reported estimation to be easier than validation, they also needed more time to complete the task. One possible reason for the longer response time for estimation is the interactivity associated with the task. The interaction time might be reduced by providing the participants a draggable line instead of a slider. In future work, it would also be interesting to capture the number of readjustments of the estimated lines to compare it with the response time and the accuracy results. 
%Further research is needed to verify this theory.

In both estimation and validation, the critical points are well above the 95\% confidence interval, meaning that people are not able to perceive regression models with an acceptable level of accuracy by statistical standards. Therefore, individuals cannot rely on their visual validation ability.
Given the relatively low self-reported expertise of the participants, the results might be different for people familiar with regression concepts, and an additional experiment solely with domain experts would be interesting for future work.

%Notably, validation and estimation were unbiased with respect to the direction of trends. 

Our study design allows us to compare our results for the perception of linear trends with those of Braun et al. for the perception of the average value~\cite{Braun.2023}. In general, both visual validation and estimation are less accurate for linear trends than for average values. This is probably due to the increased difficulty of the task, as the average value only examines the characteristic of one variable, while the linear trend examines the relationship between two variables. Nevertheless, the relation between validation and estimation remains similar, but only with respect to ODR regression. The reason for this is most likely that the vertical regression is both statistically and visually decisive when calculating and perceiving the average value. 
%One major difference in the results is that the perception of a linear trend is not symmetric, i.e., it matters whether the displayed line has a positive or negative slope deviation, whereas that doesn't matter for the average value.

The results of the experiment showed that the participants in our study were biased towards trend lines that were ``too steep''. In the estimation task, they consistently drew lines with slopes that are too high, while in the validation task they accepted significantly more lines that were too steep than lines that were too flat with respect to the true trend line. 
Possible reasons, such as confirmation bias (i.e., given a trend line, participants may identify points that support the trend line as a valid one)~\cite{Dimara.2020} or the overestimation of large values and underestimation of small values~\cite{McColeman.2021} could not be verified by our study data and therefore require further investigation.
However, the bias is reduced if the results are considered in relation to the ODR instead of the OLS regression. This indicates that, without any context or other assistance, people intuitively estimate and validate an orthogonal regression (ODR) instead of a vertical one (OLS) in scatterplots. That means, people naturally perceive errors in both variables even though they are not present in the data. As OLS is the more commonly used model for regression, we conducted a second experiment that uses additional visual augmentations to the scatterplots to improve the validation quality for OLS regression (see next section).

%Without any context or other assistance, people intuitively estimate and validate an orthogonal regression (ODR) instead of a vertical one (OLS) in scatterplots. This means, people naturally perceive errors in both variables even though they are not present in the data. This can be seen by the fact that the participants in our study were biased towards trend lines that were ``too steep''. In the estimation task, they consistently drew lines with slopes that are too high, while in the validation task they accepted significantly more lines that were too steep than lines that were too flat with respect to the true trend line. However, this bias is reduced if the results are considered in relation to the ODR instead of the OLS regression. As OLS is the more commonly used model for regression, we conducted a second experiment that uses additional visual augmentations to the scatterplots to improve the validation quality for OLS regression (see next section).

%% file: text/sec-exp2.tex
\section{Experiment 2: Visual Validation with Visual Designs}
\label{sec:exp2}

The results from experiment 1 showed that visual validation is less accurate than visual estimation. Moreover, people more likely assess orthogonal distances between data points and trend line (i.e. ODR) instead of vertical distances (i.e., OLS). However, since OLS is the more commonly-used regression model, in this section, we investigate whether the addition of visualization designs may help improve the participants' accuracy when performing visual validation with OLS regression.

Experiment~2 evaluated three common visual augmentations for regression visualization (error lines, bounding boxes, and confidence intervals), as shown in~\cref{fig:designs}. As described later in \cref{sec:designs}, we expect error lines to have the most influence on the results. Thus, our hypotheses are:
%This visual support only makes sense for validation, as the visual support for estimation would simply be to show the respective model. Thus, the experiment involved only validation task.

%\begin{itemize}[noitemsep, nolistsep]
\begin{itemize}[itemsep=0pt, parsep=3pt, leftmargin=*]
    \item \textbf{H1}: \textit{Visual designs improve the validation of OLS.} It means that people accept more statistically valid and reject more invalid trend lines with respect to OLS regression.
    \item \textbf{H2}: \textit{The visual design using error lines removes bias towards higher slopes.} This means, there is no perceptual difference between positive and negative slope deviations with respect to OLS regression anymore.
    \item \textbf{H3}: \textit{Error lines reduce the time and difficulty of the task with respect to the other designs including the unaugmented chart.}
\end{itemize}

\subsection{Visual Designs}
\label{sec:designs}

%In the context of information visualization, visual designs refer to the perceptual attributes and characteristics of graphical elements used to represent data. These designs play a crucial role in conveying information effectively to the viewer~\cite{Ware.2019}. 
Visual designs for regression validation mean additional graphical elements that are shown to the user in addition to the trend line in the visualization. We consider common visual designs for showing regression results found in literature (e.g.,~\cite{Padilla.2021, Padilla.2022,Yang.2019}) and the strategies used by participants in experiment~1 (see \cref{sec:strategies}). 
%Note, as for validation task, we focus only on designs that depend on the trend line shown, since purely data-dependent visual designs (e.g., a convex hull of the data points) would also affect the visual estimation. \remco{I don't understand this last sentence. Delete?}

\paragraph{Error Lines}

The first visual design employs error lines~\cite{Yang.2019}.
Error lines show the the vertical distance of each data point from the displayed line, emphasizing the error minimized by OLS regression (see \cref{fig:design_error}).
This potentially relieves the user of one step of the regression calculation, which should reduce participants' response times. Moreover, by visually guiding towards OLS regression, the explicit error lines should make people perceive less the ODR regression. Therefore, it should also reduce the bias toward lines that are too steep. 

\paragraph{Bounding Box}

The concept of bounding box covering the data was mentioned by several participants in the self-reported strategies and is inspired by Yang et al.~\cite{Yang.2019}. Our implementation constructs the box by moving two lines parallel to the shown line outward until they reach the outermost points. The surface is then colored with an alpha value of 0.1 for a lower opacity (see \cref{fig:design_box}). 
The resulting box highlights the slope of the line shown. This enlarged area, which includes all data points, is intended to help participants compare the slope of the line shown with the true trend of the data as a whole.

\paragraph{Confidence Interval}

Visualizing the confidence interval of a regression model is common in several different areas and applications, such as pandemic infection projections or weather forecasts~\cite{Padilla.2023, Zhang.2021, Ruginski.2016}. It shows the uncertainty in the underlying model~\cite{Padilla.2021, Padilla.2022}. All models within the confidence interval should be considered valid in a statistical sense. For our stimuli, we rotate the 95\% confidence interval of the true OLS regression model in the same way as the shown line (see \cref{fig:design_conf}). The goal is to enhance the perception of slope deviation of a line by showing all statistically valid models that accordingly have an even greater deviation.

\subsection{Experimental Setting and Participants}

We conducted a second user study on Limesurvey~\cite{LimeSurvey.2024} (with participants from Prolific~\cite{Prolific}) using the identical data and study procedure as in experiment~1. The between-subject groups were now defined by the three designs instead of the two tasks in the first study. All participants performed the visual validation task.

After filtering for attention checks, a total of 108 participants were included in the analysis (32 for \textit{error lines}, 38 for \textit{bounding box}, and 38 for \textit{confidence interval}). The demographic characteristics of the participants were similar to experiment~1: 87\% were between 20 and 40 years old with 47\% females and 52\% males. The education levels were also similar: 43\% of the participants had a bachelor's degree and 82\% responded with an expertise between 1 and 3 on the Likert scale.

Participants were compensated with \pounds3.54. The average study completion time was 17 minutes.

\subsection{Results}

We can directly compare the results for the visual designs with the \textit{base line} (i.e., lines without additional augmentations) from experiment~1.

\paragraph{Accuracy}

%The acceptance rates of each visual design, summarized by positive and negative trends and positive and negative slope deviation, are shown in \cref{fig:designs_comp}. 

\begin{figure}%[tbh!]
    \centering
    \includegraphics[width=0.7\columnwidth]{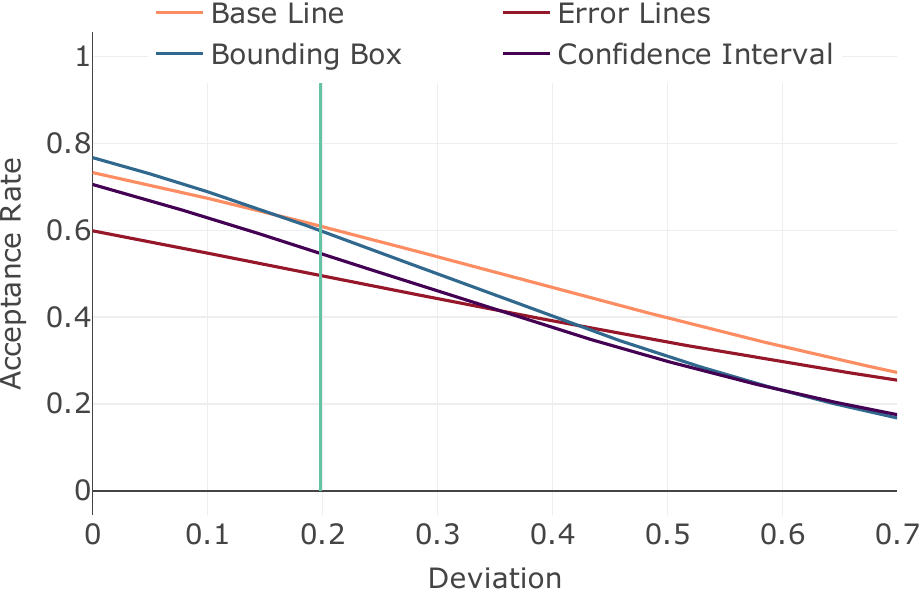}
    \caption{Comparison of validation accuracy (absolute deviation) with respect to OLS regression for the visual designs. Green line: Statistical 95\% CI.}
    \label{fig:designs_comp_ols}
    
     % \begin{subfigure}[t]{0.7\columnwidth}
     %     \centering
     %     \includegraphics[width=\columnwidth]{features_comp_odr}
     %     \caption{ODR regression.}
     %     \label{fig:designs_comp_odr}
     % \end{subfigure}
     %    \caption{Comparison of validation accuracy (absolute deviation) with respect to both regression models for the visual designs. Green line: Statistical 95\% CI.}
     %    \label{fig:designs_comp}
\end{figure}

With respect to OLS regression (see \cref{fig:designs_comp_ols}), the pairwise KS-test indicated significant differences in the acceptance rates between all designs (\cref{tab:accuracy}).
%was found using the Kruskal-Wallis test ($p\ll0.01$). The pairwise Wilcoxon test showed that the acceptance rates for the \textit{base line} are significantly different from the lines with visual designs (\cref{tab:accuracy}). There are no significant differences between the visual designs. 

For the \textit{bounding box} and the \textit{confidence interval}, a similar number of valid lines (i.e., trend lines within the CI) were accepted compared to the \textit{base line} condition, while a slightly higher amount of invalid lines were rejected. For the \textit{error lines}, fewer lines were accepted that were within the statistical 95\% confidence interval. The true trend lines were accepted only 60\% of the time.

The overall decrease in the acceptance rates with visual designs have lowered the critical point between acceptance and rejection of a shown line for all three designs ($crit_{error} = 0.190$, $crit_{box} = 0.251$, $crit_{conf} = 0.298$). 

In sum, people rejected slightly more invalid models with visual designs, but did not accept more valid models. Therefore, our hypothesis \textbf{H1} cannot be rejected.

\begin{table}[b]
    \centering
    \begin{tabular}[\columnwidth]{l|c|c|c|c}
        \textbf{p-value} & \textit{base} & \textit{error} & \textit{box} & \textit{conf} \\
        \hline
        \textit{base} & - &  &  &  \\
        \hline
        \textit{error} &  $\ll0.01$ & - &  &   \\  
        \hline
        \textit{box} & $\ll0.01$ & $\ll0.01$ & - &   \\
        \hline
        \textit{conf} & $\ll0.01$ & $\ll0.01$ & $<0.01$ & -  \\
    \end{tabular}
    \caption{p-values of the pairwise KS-test for the analysis of the acceptance rates of OLS regression.}
    \label{tab:accuracy}
\end{table}

%With respect to ODR regression (see~\cref{fig:designs_comp_odr}), the results for the \textit{confidence interval} behave similar to those of the \textit{base line} by increasing the acceptance rate of valid trend lines and decreasing it for invalid ones. In contrast, the acceptance rates with the \textit{bounding box} and the \textit{error lines} were almost the same as in the OLS regression (see~\cref{fig:designs_comp_ols}). That indicates that the addition of bounding boxes and error lines in the visualization guides people away from considering trend lines as ODR regressions and toward OLS regressions.

\paragraph{Bias}

\begin{figure}%[tbh!]
    \centering
    \begin{subfigure}[t]{\columnwidth}
         \centering
         \includegraphics[width=\columnwidth]{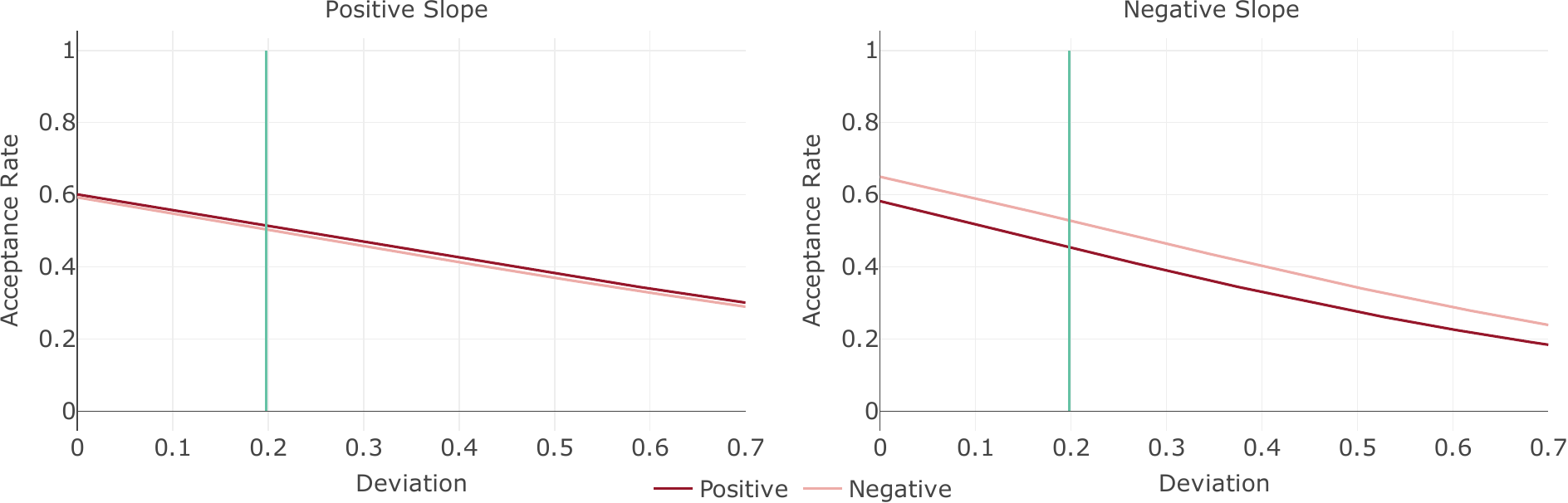}
         \caption{Error Lines.}
         \label{fig:error_error_comp}
     \end{subfigure}
    
     \begin{subfigure}[t]{\columnwidth}
         \centering
         \includegraphics[width=\columnwidth]{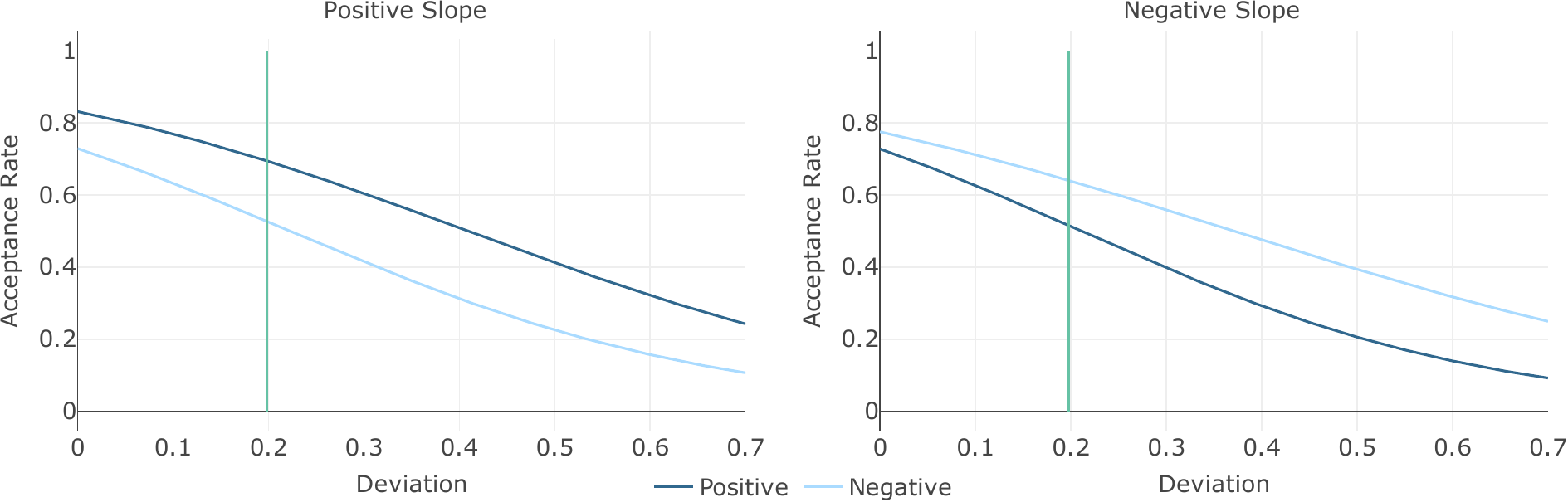}
         \caption{Bounding Box.}
         \label{fig:box_error_comp}
     \end{subfigure}

     \begin{subfigure}[t]{\columnwidth}
         \centering
         \includegraphics[width=\columnwidth]{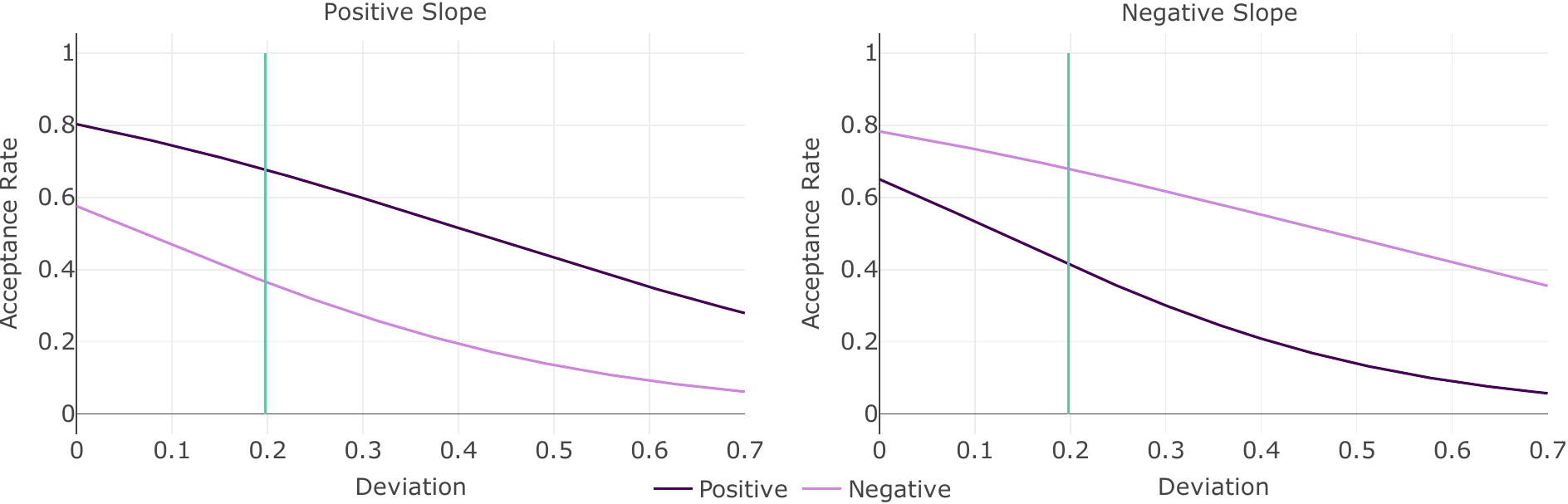}
         \caption{Confidence Interval.}
         \label{fig:conf_error_comp}
     \end{subfigure}
        \caption{Comparison of the validation acceptance rates for the different visual designs for positive and negative deviations and for positive and negative trends with respect to OLS regression. Green line: statistical 95\% CI.}
        \label{fig:designs_error_comp}
\end{figure}

\cref{fig:designs_error_comp} shows the differentiated acceptance rates for the designs.
For the \textit{error lines}, the deviation difference between the acceptance' critical points of positive and negative slope deviation decreased ($\Delta^{pos}_{error}=0.020$ for positive trends and $\Delta^{neg}_{error}=0.117$ for negative trends). While the acceptance rates for negative trends still differ (KS: $p\ll0.01$, $D=0.182$), they are not significantly different for positive trends anymore (KS: $p>0.71$, $D=0.044$).
This result suggests that the perception of OLS regression was improved for the participants with \textit{error lines}. It partially supports hypothesis \textbf{H2}.

Similar to the \textit{base line}, the \textit{confidence interval} biased the perception of positive and negative slope deviation ($\Delta^{pos}_{conf}=0.348$, $\Delta^{neg}_{conf}=0.353$). The \textit{bounding box} showed a slight improvement in the perception of lines that were too steep ($\Delta^{pos}_{box}=0.186$, $\Delta^{neg}_{box}=0.159$).

%Comparing the results for positive and negative trends, visual validation is still unbiased with visual designs.

\paragraph{Response Time and Difficulty}

\begin{figure}
    \centering
    \begin{subfigure}[t]{0.7\columnwidth}
         \centering
         \includegraphics[width=\columnwidth]{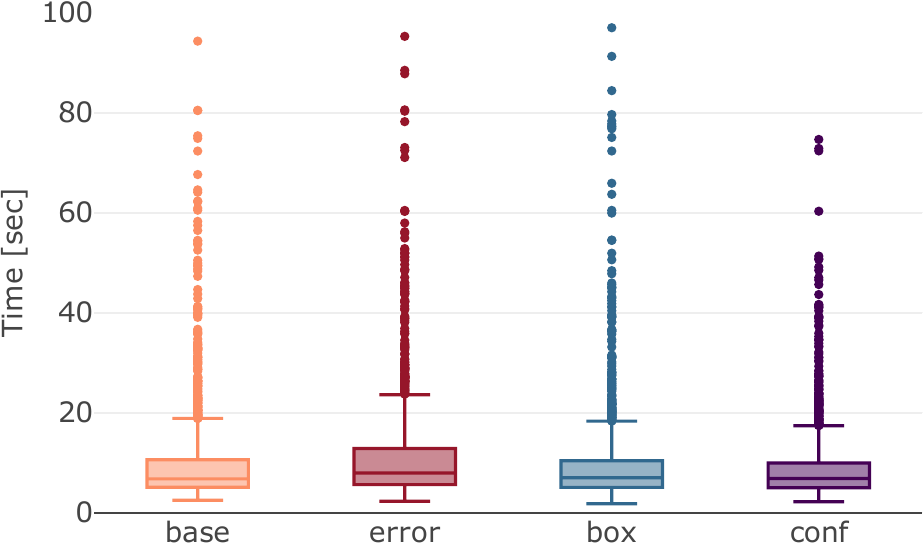}
         \caption{Response times.}
         \label{fig:exp2_time}
     \end{subfigure}
    
     \begin{subfigure}[t]{0.7\columnwidth}
         \centering
         \includegraphics[width=\columnwidth]{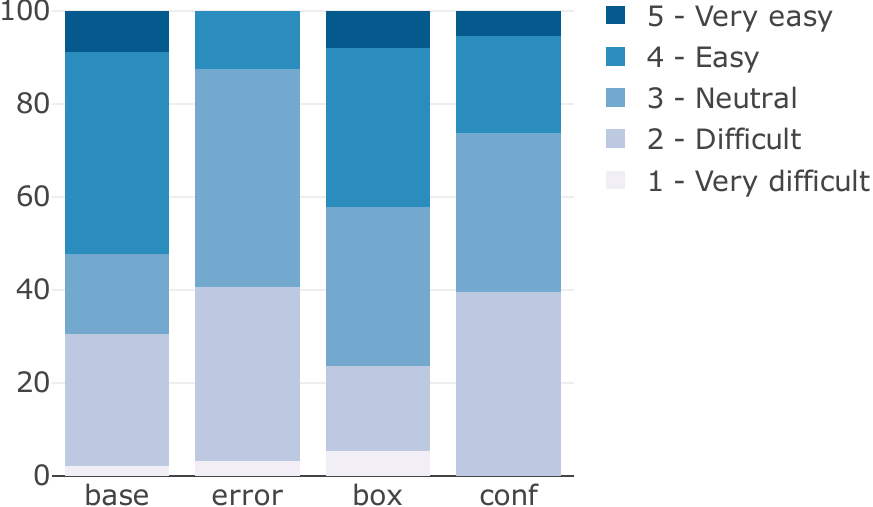}
         \caption{Self-reported task difficulty.}
         \label{fig:exp2_diff}
     \end{subfigure}
        \caption{Response times and self-reported task difficulty for the different visual designs in experiment 2.}
        \label{fig:exp2_time_diff}
\end{figure}

The response times (measured in seconds) and participants' self-reported task difficulty on a 5-point Likert scale are shown in \cref{fig:exp2_time_diff}. 

The Kruskal-Wallis test indicated a significant difference in the response times ($p\ll0.01$, $\chi^2=112.42$). A pairwise Wilcoxon test of the visual designs (\cref{tab:time}) showed that the response times with the \textit{error lines} were significantly higher than for all other designs ($\mu_{error}=12.64sec$). Among the other designs, the \textit{confidence interval} had the fastest response times ($\mu_{conf}=9.21sec$), followed by the \textit{bounding box} ($\mu_{box}=9.96sec$) and the \textit{base line} ($\mu_{base}=10.85sec$).

A significant difference in the designs was also found in the perceived difficulty (Chi-squared: $p<0.05$, $\chi^2=21.19$). The task was perceived to be the easiest with the \textit{base line} ($\mu_{base}=3.28$). The only significance could be found when comparing the \textit{base line} to the \textit{error lines} (\cref{tab:diff}) where the \textit{error lines} was perceived as the most difficult ($\mu_{error}=2.69$). In between the \textit{base line} and the \textit{error lines} were the \textit{bounding box} ($\mu_{box}=3.21$) and the \textit{confidence interval} ($\mu_{conf}=2.92$).

Summarizing the results, hypothesis \textbf{H3} is not supported.

\begin{table}[tb]
    \begin{subtable}[tb]{\columnwidth}
        \centering
        \begin{tabular}{l|c|c|c|c}
            \textbf{p-value} & \textit{base} & \textit{error} & \textit{box} & \textit{conf} \\
            \hline
            \textit{base} & - &  &  &  \\
            \hline
            \textit{error} &  $\ll0.01$ & - &  &   \\  
            \hline
            \textit{box} & \color{Grey}{$1.00$} & $\ll0.01$ & - &   \\
            \hline
            \textit{conf} & \color{Grey}{$0.92$} & $\ll0.01$ & \color{Grey}{$0.58$} & -  \\
        \end{tabular}
    \caption{Response times.}
    \label{tab:time}
    \end{subtable}
    
    \begin{subtable}[tb]{\columnwidth}
        \centering
        \begin{tabular}{l|c|c|c|c}
            \textbf{p-value} & \textit{base} & \textit{error} & \textit{box} & \textit{conf} \\
            \hline
            \textit{base} & - &  &  &  \\
            \hline
            \textit{error} &  $<0.01$ & - &  &   \\  
            \hline
            \textit{box} & \color{Grey}{$0.38$} & \color{Grey}{$0.06$} & - &   \\
            \hline
            \textit{conf} & \color{Grey}{$0.11$} & \color{Grey}{$0.36$} & \color{Grey}{$0.18$} & -  \\
        \end{tabular}
    \caption{Self-reported difficulties.}
    \label{tab:diff}
    \end{subtable}
\subfigsCaption{p-values of the pairwise Wilcoxon-test for the response time analysis and pairwise chi-squared test for analysis of the self-reported difficulties.}
\label{tab:timer}
\end{table}

\subsection{Discussion}

The results of experiment 2 show that people consistently reject trend lines more often when they are presented with visual designs regardless of whether the trend lines are valid or invalid. There could be several different reasons for this: 1) The visual designs help to identify incorrect regression models. 2) People are generally more skeptical when shown model results with visual designs because they are not used to or do not understand this type of presentation. 3) People perceive the type of model to not fit to the underlying data. In our study, we intentionally did not provide an explanation for the designs to allow for a purely perceptual study. It is therefore possible that a prior explanation of the visual designs would improve participants' understanding and thus the results. Similarly, the results may be influenced by the low statistical expertise of our participants.

With respect to OLS regression, the acceptance thresholds improved with all designs compared with the \textit{base line}. This is due to slightly higher rejection rates of invalid trend lines. The validation accuracy of valid lines did not improve with any design.

As with the \textit{base line}, there is no bias in trend direction with the visual designs. 
Although confidence intervals are commonly used to represent statistical uncertainty, the bias in slope deviation is greatest with the addition of the confidence interval in the visualization. 
The error lines, on the other hand, provided an unbiased validation in terms of ``too steep'' and ``too flat'' trend lines.

%For visual validation of ODR regression, people are most accurate with the \textit{base line}. With the \textit{bounding box} and \textit{error lines} there is almost no change in the acceptance rates compared to OLS regression. Besides the base line, the results match ODR regression the most with the confidence interval. Since the accuracy didn't improve with the confidence interval, it should be considered to not show confidence intervals to reduce visual clutter.

The addition of \textit{error lines} in the visualization should in theory reduce the cognitive effort of visual validation, because the cognitive calculation of errors is not needed. However, participants took longer to complete the validation task and found it more difficult than with the \textit{base line}, according to the results of our study. This suggests that people either did not fully understand the concept of error lines without explanation, that the processing of additional information is cognitively demanding, or that it forces people to intensify their thinking about the shown line and to correct their bias.

The results of experiment 2 showed that the addition of commonly used visual designs for visualizing regression in a scatterplot does not significantly improve people's ability to validate models visually. Therefore, we are unable to provide design guidelines. As visual estimation remains more accurate, this suggests that guiding people to cognitive estimation as a first step of visual validation might improve accuracy.
%With the results discussed we provide the following \textbf{design guidelines} for the use of visual designs for visual validation:
%\begin{itemize}[noitemsep, nolistsep]
%    \item Model result should be presented with bounding box if OLS regression is desired.
%    \item Model result should be presented without visual designs if ODR regression is desired.
%    \item Regression error lines provide unbiased visual validation.
%    \item Showing the regression confidence interval doesn't support people in visual validation.
%\end{itemize}

%% file: text/sec-futurework.tex
\section{Limitations and Future Work}
\label{sec:futurework}

We investigated the visual estimation and validation of linear trend lines in scatterplots. The findings and limitations in our experiments may suggest new research questions and future directions.

\smallskip\noindent \textbf{Model complexity:} We found that individuals' ability to visually validate a linear model is lower than their ability to validate a constant model (i.e., averages~\cite{Braun.2023}).
This raises the question whether the ability to visually validate is dependent on the complexity of a model. In order to answer this question, however, it would be necessary to define model complexity in relation to visual perception.

\smallskip\noindent \textbf{Data characteristics - Outliers:} Our study analyzed data with normal distribution as assumed by OLS regression models. However, real world datasets may have special characteristics that impact the regression. For example, Correll and Heer~\cite{Correll.2017} studied the influence of noise and outlier in the data. The addition of outliers would shift the validation question of a correct or incorrect model to a question of including or excluding the outliers in the regression. This question about the correctness of the model itself could be extended to a study on the validation of model types. In this, people would have to be decided whether the type of model fits the data or not. In this paper, we examined the fit of the parameters of a fixed model.

\smallskip\noindent \textbf{Visual designs for model validation:} We tested four designs (unmodified base line and three visual design augmentations) that are commonly used in visualizing regression models. The addition of the three visual designs in experiment~2 failed to improve both the accuracy and bias of visual validation. It would be interesting to see whether the results improve in a separate study where explanation and context to the data and the visual designs are made available to the participant. Additional think-aloud sessions could further provide insights into people's visual validation process and the use of visual designs. Altogether, the development of novel visual designs for model validation that improves both accuracy and mitigates bias is a future challenge.

%% file: text/sec-conclusion.tex
\section{Conclusion}
\label{sec:conclusion}

In summary, our research examines the effectiveness of visual validation in assessing linear regression models shown in scatterplots. We conducted two empirical experiments to gain insight into individuals' abilities to visually validate linear trends and the impact of common visualization designs on validation quality. 

The first experiment showed that participants were more accurate at visually estimating linear trends in scatterplots than at visually validating them. In addition, our results revealed a bias in slope deviation (i.e., toward slopes that are "too steep"), but no bias in trend direction. Additional analysis provides evidence that people naturally assess orthogonal regression (ODR) rather than the most commonly used vertical regression (OLS). This indicates that people assume errors in both variables rather than in just the y-coordinate.

The second experiment aimed to evaluate whether incorporating common visualization designs such as error lines, bounding boxes, and confidence intervals could improve visual validation. Despite the reduction in validation bias observed with error lines, none of the tested designs yielded the desired improvements in accuracy. 

Our results emphasize the limitations of relying solely on visual model validation for linear regression models in scatterplots. Further research is needed to investigate the underlying cognitive processes involved in visual validation tasks in order to find appropriate visual solutions for supporting visual model validation.